\documentclass{article}

\usepackage{longcat_style}
\settitlelogo{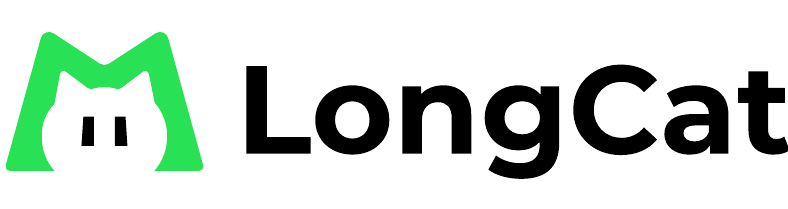}
\settitlerightlogo{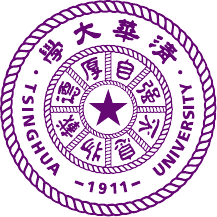}
\renewcommand{\headeright}{Preprint}
\renewcommand{\shorttitle}{%
  \mbox{\normalfont\fontsize{8}{9.5}\selectfont RAP: A Dataset for Paper-to-Poster Generation via Declarative Markup from Human-Crafted Exemplars}%
}

\usepackage{iftex}
\ifPDFTeX
  \usepackage[utf8]{inputenc}
\else
  
\fi
\usepackage[T1]{fontenc}
\usepackage{booktabs,nicefrac,microtype}
\usepackage{xcolor,graphicx}
\usepackage{amsmath,amssymb,amsfonts}
\usepackage{multirow,enumitem,tabularx,makecell,pifont}
\usepackage{subcaption,colortbl,xspace,float}
\usepackage[numbers,sort&compress]{natbib}
\usepackage{url}
\usepackage{placeins}
\usepackage{hyperref}
\usepackage{fontawesome5}
\usepackage{tcolorbox}
\hypersetup{
  colorlinks=true,
  linkcolor=blue,
  urlcolor=blue,
  citecolor=blue,
  pdftitle={RAP: A Dataset for Paper-to-Poster Generation via Declarative Markup from Human-Crafted Exemplars},
  pdfauthor={Haonan Han, Yukang Lin, Minliang Lin, Haoxian Tan, Jun-Yan He, Rui Yang, Zhi-Qi Cheng, Jie Hu, Guanglu Wan, Xiu Li}
}
\newcommand{\cmark}{\ding{51}}
\newcommand{\xmark}{\ding{55}}
\definecolor{darkgreen}{RGB}{0,200,0}
\definecolor{purple}{RGB}{153,51,204}
\definecolor{blue}{RGB}{37,99,235}
\definecolor{lightaccentblue}{RGB}{219,234,254}
\definecolor{lightaccentgreen}{HTML}{64C9D2}
\definecolor{azure}{rgb}{0.94,0.97,1.0}
\definecolor{lavender}{rgb}{0.96,0.94,1.0}

\newcommand{\DatasetName}{\textit{RAP}\xspace}
\newcommand{\DatasetNameFull}{RAP (Reconstruct A Poster)\xspace}
\newcommand{\MethodName}{PosterRAG\xspace}

\definecolor{badgehome}{HTML}{2BE155}
\definecolor{badgehomeink}{HTML}{FFFFFF}
\definecolor{badgegithub}{HTML}{24292F}
\definecolor{badgemodel}{HTML}{7C3AED}
\definecolor{badgemodelink}{HTML}{FFFFFF}
\definecolor{badgedemo}{HTML}{FF7A00}
\definecolor{badgedemoink}{HTML}{FFFFFF}
\definecolor{badgedataset}{HTML}{2563EB}
\definecolor{badgedatasetink}{HTML}{FFFFFF}

\newtcbox{\linkbadgebox}[1][badgehome]{
  on line,
  arc=8pt,
  boxrule=0.4pt,
  colback=#1,
  colframe=#1!82!black,
  boxsep=1pt,
  left=5pt,
  right=5pt,
  top=2pt,
  bottom=2pt,
  tcbox raise base,
  nobeforeafter
}

\newcommand{\linkbadge}[4]{%
  \begingroup
  \fontsize{9}{10.5}\selectfont
  \if\relax\detokenize{#1}\relax
    \linkbadgebox[#2]{\makebox[6.6em][c]{\rule[-0.25em]{0pt}{1.25em}\textcolor{#3}{#4}}}%
  \else
    \href{#1}{\linkbadgebox[#2]{\makebox[6.6em][c]{\rule[-0.25em]{0pt}{1.25em}\textcolor{#3}{#4}}}}%
  \fi
  \endgroup
}

\newcommand{\homepage}[1]{%
  \linkbadge{#1}{badgehome}{white}{\textcolor{badgehomeink}{\faGlobe}~\textbf{Homepage}}%
}

\newcommand{\github}[1]{%
  \linkbadge{#1}{badgegithub}{white}{\faGithub~\textbf{Github}}%
}

\newcommand{\huggingface}[1]{%
  \linkbadge{#1}{badgemodel}{badgemodelink}{%
    \raisebox{-0.12em}{\includegraphics[height=1em]{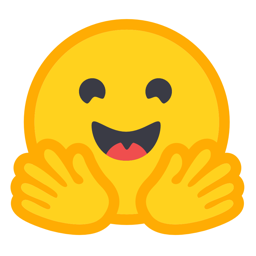}}~\textbf{Model}%
  }%
}

\newcommand{\demobadge}[1]{%
  \linkbadge{#1}{badgedemo}{badgedemoink}{%
    \raisebox{-0.12em}{\includegraphics[height=1em]{figures/icon/huggingface.png}}~\textbf{Demo}%
  }%
}

\newcommand{\datasetbadge}[1]{%
  \linkbadge{#1}{badgedataset}{badgedatasetink}{%
    \raisebox{-0.12em}{\includegraphics[height=1em]{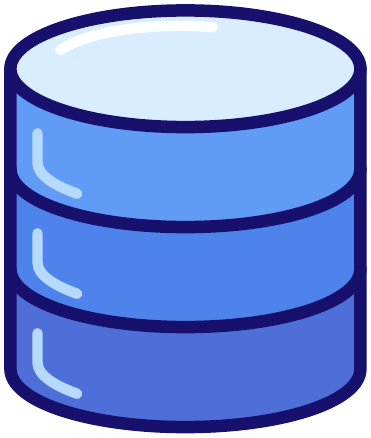}}~\textbf{Dataset}%
  }%
}

\title{\textit{RAP}: A Dataset for Paper-to-Poster Generation via Declarative Markup from Human-Crafted Exemplars}
\author{%
  \bfseries Haonan Han$^{1,2}$ \quad Yukang Lin$^{1}$ \quad Minliang Lin$^{2}$ \quad
  Haoxian Tan$^{2}$ \quad Jun-Yan He$^{2}$\thanks{Project lead.} \quad Rui Yang$^{3}$ \\[3pt]
  \bfseries Zhi-Qi Cheng$^{4}$ \quad Jie Hu$^{2}$ \quad
  Guanglu Wan$^{2}$ \quad Xiu Li$^{1}$\thanks{Corresponding author.} \\[6pt]
  \normalfont\small $^{1}$Tsinghua University \quad $^{2}$Meituan Longcat \quad $^{3}$The University of Hong Kong \quad $^{4}$University of Washington \\[7pt]
  \normalfont\normalsize
  \mbox{%
    \homepage{https://reconstruct-a-poster.github.io/}\hspace{6pt}%
    \github{https://github.com/VincentHancoder/RAP}\hspace{6pt}%
    \huggingface{https://huggingface.co/VincentHancoder/RAP-Qwen3.8-27B-SFT-RL}\hspace{6pt}%
    \demobadge{https://huggingface.co/spaces/VincentHancoder/RAP}\hspace{6pt}%
    \datasetbadge{https://huggingface.co/datasets/VincentHancoder/RAP}%
  }
}

\begin{document}
\raggedbottom
\maketitle

\begin{figure}[H]
\centering
\includegraphics[width=\textwidth]{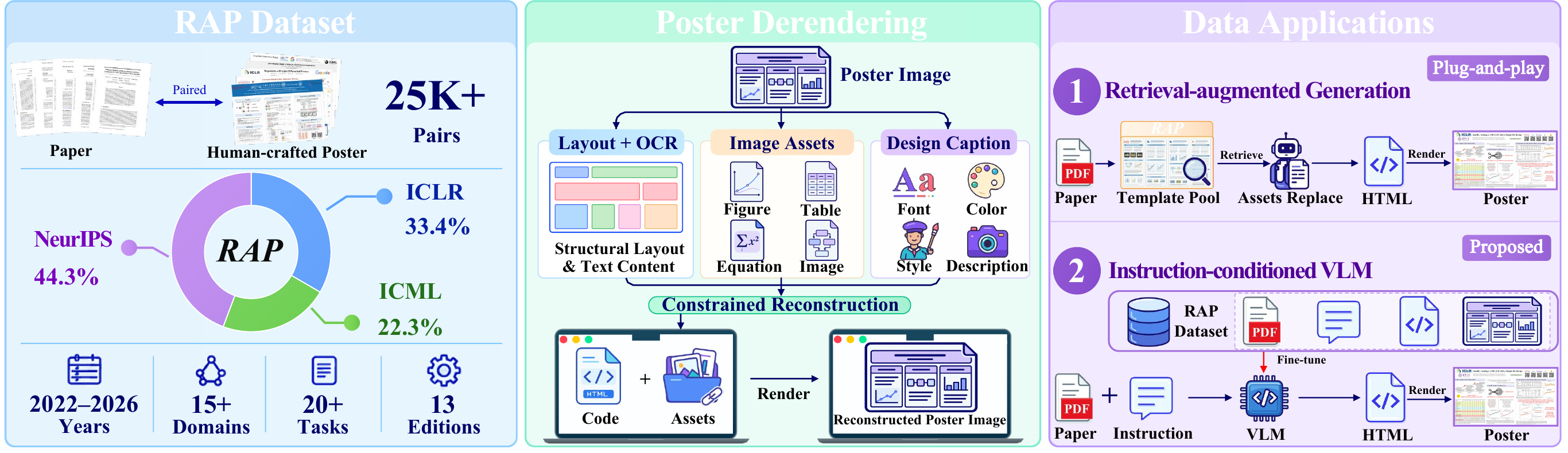}
\caption{Overview of \DatasetName. \textbf{Left:} paired scientific papers and human-crafted posters from major machine learning conferences. \textbf{Middle:} automated derendering reconstructs posters as editable HTML with linked assets. \textbf{Right:} the resulting data support retrieval-augmented template adaptation and instruction-conditioned VLM post-training for HTML-based poster generation.}
\label{fig:tearser}
\end{figure}

\begin{abstract}
\fontsize{10}{12}\selectfont
Making a conference poster is visual coding in miniature: an entire paper must be compressed onto a single page, which exercises summarization, visual design, and code implementation at once, using nothing beyond the paper itself. Yet benchmarks for this task in the context of visual coding remain scarce: existing resources are small and annotated only with layout boxes, so evaluation must fall back on VLM-as-judge or human checklists rather than objective render-and-compare, and the generation methods they support design layouts from scratch, leaving human-crafted posters unusable as either a retrievable design library or as verified supervision. We introduce \textbf{\DatasetName{}} (Reconstruct A Poster), a dataset and benchmark for paper-to-poster generation as a visual coding task. \DatasetName{} contains 25K+ verified paper-poster pairs from various machine learning conferences held between 2022 and 2026, annotated with a three-level content taxonomy, plus a held-out benchmark of 150 test papers each paired with its human poster. A fully automated derendering pipeline turns each poster into styled HTML/CSS of typed, refillable slots, so every sample is a paper-poster-markup triplet verifiable by rendering against the source poster. The corpus serves both inference and training: \MethodName, a training-free multi-agent pipeline, retrieves an exemplar and refills its slots without touching layout or styling, while an SFT+RL recipe trains a dedicated model from the same triplets. An online demo is available at \url{https://reconstruct-a-poster.github.io/}.

\end{abstract}

\clearpage
\section{Introduction}
\label{sec:introduction}

Automatically condensing a scientific paper into a conference poster is an ideal yet long-overlooked validation task for visual coding. The task exercises three capabilities that are individually measurable---summarization, layout and visual design, and code implementation: a model must extract the salient content of a full-length paper, organize that content into a clear layout with coherent visual design under limited space and a fixed aspect ratio, and render the result legibly for human readers. All three draw solely on information that the paper itself supplies, so every model operates on identical inputs, free from the variance introduced by web search.

However, existing benchmarks remain limited in scale, annotation depth, and utilization. \textbf{First, scale and verifiability.} Paper2Poster provides 100 paper-poster pairs~\citep{posteragent2025} and P2PEval 121~\citep{p2p2025}. Larger collections supply abstracts~\citep{saxena2025postersum} or full papers for only 100 of 7{,}855 posters~\citep{tanaka2024scipostlayout}. SciPostGen pairs 18{,}097 papers and posters but targets layout-level annotation~\citep{inadumi2025scipostgen}. Generation benchmarks rely on subjective, costly VLM judgments or human checklists~\citep{posteragent2025, p2p2025} rather than direct render-and-compare verification, limiting reproducibility and statistical robustness.
\textbf{Second, task formulation and annotation depth.} Resources annotate boxes, categories, reading order, and hierarchy~\citep{tanaka2024scipostlayout, tanaka2025scipostlayouttree}, but lack renderable markup reconstructing source posters for end-to-end visual coding.
\textbf{Third, utilization.} Methods largely generate layouts from scratch, while a template-based system retrieves from a few curated slides~\citep{zheng2025pptagent}. P2PInstruct supplies model-generated rather than human-grounded HTML designs~\citep{p2p2025}. Existing resources lack reusable human-crafted design libraries and verified paper-poster supervision for generator training. Table~\ref{tab:dataset-comparison} summarizes these gaps.

\begin{table}[H]
\centering
\caption{Comparison of representative scientific poster datasets and benchmarks. Our resource links full papers and human-crafted poster exemplars to reconstructed, styled HTML with typed, refillable content slots. \textsuperscript{*}P2PInstruct counts instruction--response pairs with synthetic instruction/HTML targets; human-crafted posters serve as generation references. \textsuperscript{\ddag}Full-paper coverage is partial: 100 posters in SciPostLayout; SciPostLayoutTree inherits partial coverage through SciPostLayout.}
\label{tab:dataset-comparison}
\scriptsize
\setlength{\tabcolsep}{2pt}
\resizebox{\textwidth}{!}{%
\begin{tabular}{@{}l  r c c c l@{}}
\toprule
\textbf{Dataset} & \textbf{Scale} & \makecell{\textbf{Full}\\\textbf{Paper}} & \makecell{\textbf{Renderable}\\\textbf{Markup}} & \makecell{\textbf{Human-crafted}\\\textbf{Poster}} & \textbf{Annotations} \\
\midrule
SciPostLayout~\citep{tanaka2024scipostlayout} & 7{,}855 & \textcolor{darkgreen}{\cmark}\textsuperscript{\ddag} & \textcolor{red}{\xmark} & \textcolor{darkgreen}{\cmark} & Boxes \\
PosterSum~\citep{saxena2025postersum} & 16{,}305 & \textcolor{red}{\xmark} & \textcolor{red}{\xmark} & \textcolor{darkgreen}{\cmark} & Abstracts, topic labels \\
Paper2Poster~\citep{posteragent2025} & 100 & \textcolor{darkgreen}{\cmark} & \textcolor{red}{\xmark} & \textcolor{darkgreen}{\cmark} & QA pairs \\
P2PEval~\citep{p2p2025} & 121 & \textcolor{darkgreen}{\cmark} & \textcolor{red}{\xmark} & \textcolor{darkgreen}{\cmark} & Checklists \\
SciPostLayoutTree~\citep{tanaka2025scipostlayouttree} & 7{,}849 & \textcolor{darkgreen}{\cmark}\textsuperscript{\ddag} & \textcolor{red}{\xmark} & \textcolor{darkgreen}{\cmark} & Boxes, OCR, reading order, hierarchy \\
SciPostGen~\citep{inadumi2025scipostgen} & 18{,}097 & \textcolor{darkgreen}{\cmark} & \textcolor{red}{\xmark} & \textcolor{darkgreen}{\cmark} & Boxes, OCR, figure/table links \\
P2PInstruct~\citep{p2p2025} & 30{,}460 & \textcolor{red}{\xmark} & \textcolor{darkgreen}{\cmark} & \textcolor{red}{\xmark}\textsuperscript{*} & Instructions, figure descriptions, text, HTML \\
\midrule
\cellcolor{lavender}\textbf{\DatasetName (Ours)} & \cellcolor{lavender}\textbf{25{,}435} & \cellcolor{lavender}\textcolor{darkgreen}{\cmark} & \cellcolor{lavender}\textcolor{darkgreen}{\cmark} & \cellcolor{lavender}\textcolor{darkgreen}{\cmark} & \cellcolor{lavender}\makecell[l]{\textbf{Boxes, OCR, HTML, figure assets,}\\\textbf{asset links, design captions, taxonomy}} \\
\bottomrule
\end{tabular}%
}
\end{table}

We therefore introduce \textbf{\DatasetNameFull}, a dataset and benchmark for paper-to-poster generation as a visual coding task. At its core is the largest paper-poster collection to date: 25{,}435 verified pairs collected from NeurIPS, ICML, and ICLR (2022--2026) via OpenReview, each enriched with paper-level metadata, a distilled introduction summary, and annotations from a three-level taxonomy of 15 domains, 20 tasks, and 39 methods that support content-aware retrieval and analysis. Complementing the corpus, we curate a held-out benchmark of 150 test papers, 10 per domain, each paired with its ground-truth human poster. Because every sample carries its own human-crafted reference, render-and-compare verification is available at the scale of the whole corpus rather than only on a handful of curated examples, making measurement objective and reproducible instead of dependent on judge-based scoring.

To demonstrate the dataset's utility at both inference time and training time, we develop two applications that exploit its complementary supervision signals.
\emph{(1) Training-free retrieval-augmented generation.} We present \textbf{\MethodName}, a multi-agent pipeline that treats the dataset as a reusable design library. A hierarchical retrieval cascade---categorical filtering, narrative matching, and aesthetic scoring---selects the best-matching exemplar for an input paper; a mapping agent then programmatically refills the retrieved template's typed slots with paper content while leaving all layout and styling untouched; and a VLM reviewer iteratively repairs residual visual defects. Because \MethodName inherits human-crafted designs unmodified rather than hallucinating layouts, it achieves state-of-the-art diversity and distributional human-likeness against strong baselines.
\emph{(2) Post-training a dedicated paper-to-poster model.} The triplets likewise provide direct input--output supervision: we apply supervised fine-tuning (SFT) to a multimodal model on paper--markup pairs, then further optimize it with reinforcement learning using verifiable rewards~\citep{shao2024deepseekmath}.

Our contributions are as follows:
\begin{itemize}[leftmargin=*,noitemsep]
  \item \textbf{A large-scale dataset and benchmark.} \DatasetName{} is the largest paper-poster resource to date, comprising 25{,}435 verified pairs with three-level content annotations and a held-out benchmark of 150 test papers. Because every sample carries a human-crafted reference, evaluation proceeds by objective render-and-compare rather than by judge-based scoring alone.

  \item \textbf{An automated derendering pipeline.} A fully automated pipeline converts raster posters into styled, renderable HTML/CSS in which every element is an addressable, typed, and refillable slot. 

  \item \textbf{Two dataset-driven applications.} \MethodName{}, a training-free retrieval-augmented pipeline, and an SFT+RL recipe for training a dedicated paper-to-poster model demonstrate that the same corpus serves as a reusable design library at inference time and as verified supervision at training time.
\end{itemize}

\section{Related Work}
\label{sec:related}

\paragraph{Visual Coding Data and Benchmarks.}
Visual coding has emerged as a rigorous testbed for multimodal models. pix2code~\citep{beltramelli2018pix2code} pioneers code generation from GUI screenshots; Design2Code~\citep{si2024design2code} and Vision2Web~\citep{he2026vision2web} extend the task to real webpages and multi-page websites, the latter incorporating agent-based verification; Plot2Code~\citep{wu2024plot2code} and ChartMimic~\citep{yang2024chartmimic} target scientific plots and charts. These benchmarks, however, all overlook the paper-to-poster task: their inputs are screenshots or short specifications of a single, self-contained view, whereas poster generation requires condensing a full-length paper into a space-constrained, information-dense design and implementing it as code. \DatasetName is constructed around the requirements of visual coding: the full paper supplies the long-context input, every poster is derendered into styled, renderable HTML/CSS with typed, refillable slots, and reconstruction fidelity is measured by direct render-and-compare against the human reference, making paper-to-poster, for the first time, a visual coding benchmark whose outputs are verifiable at scale.

\paragraph{Paper-to-Poster Generation.}
Existing resources for this task supply posters and, at best, layout-level annotations. Paper2Poster~\citep{posteragent2025} and P2PEval~\citep{p2p2025} pair papers with only a few hundred raw poster images and rely on VLM judges for evaluation; PosterSum~\citep{saxena2025postersum} pairs posters with abstracts rather than full papers; SciPostLayout and SciPostLayoutTree~\citep{tanaka2024scipostlayout, tanaka2025scipostlayouttree} annotate layout boxes, reading order, and figure correspondence, and SciPostGen~\citep{inadumi2025scipostgen} does so at larger scale, but none provides renderable markup that reconstructs the poster. Approaches to generation fall into two families. \emph{Text-to-image generation.} Recent progress in visual text rendering~\citep{chen2024textdiffuser, tuo2024anytext, bizgen2025} and in frontier image generators~\citep{openai2025gptimage, nanobanana2025, seedream2025, flux2024, ideogram2025} enables the synthesis of poster-like images directly from prompts. Their outputs, however, are non-editable raster images whose text fidelity degrades sharply on dense scientific content, and they offer no guarantee that the rendered content faithfully reflects the source paper. \emph{LLM-based pipelines.} Paper2Poster~\citep{posteragent2025}, P2P~\citep{p2p2025}, and PosterGen~\citep{zhang2026postergen} decompose the task into paper parsing, content planning, and markup rendering in HTML or PPTX, typically with multi-agent checking and refinement, whereas PosterLLaVa~\citep{yang2024posterllava} and PosterLlama~\citep{yang2024posterllama} fine-tune multimodal models for the content-aware layout sub-problem in isolation. Instruction-tuning data for the task is model-generated rather than human-grounded: P2PInstruct~\citep{p2p2025} supplies HTML targets synthesized by models rather than derived from human-crafted posters. Document-to-slide systems~\citep{fu2022doc2ppt, zheng2025pptagent} follow the same parse--plan--render paradigm; in particular, PPTAgent~\citep{zheng2025pptagent} retrieves from a small set of manually curated slide templates, but its rigid filling mechanism often fails to adapt content appropriately. Both families thus confront a common trade-off: either sacrifice editability and content fidelity, or generate layouts from scratch without grounding in human design. \DatasetName addresses both sides of this trade-off: its derendered, human-crafted posters ground a retrieval-augmented pipeline that inherits professional designs rather than inventing them, and its paper-poster-markup triplets provide direct supervision for post-training a dedicated generation model with reference-based rewards.

\section{\DatasetName: Large-Scale Poster Derendering}
\label{sec:derendering}

\noindent \textbf{Data Collection and Statistics.} 
The RAP dataset is built from accepted papers at three top-tier ML conferences from 2022 through 2026. Using OpenReview, we collect raw poster files together with their associated paper records. For each submission, we retrieve three types of information: 
\textit{(i)} the paper PDF or LaTeX source when available from supplementary material; 
\textit{(ii)} the author-uploaded poster file in either PDF or image format; and 
\textit{(iii)} metadata including title, authors, abstract, venue, year, and keywords.
We then normalize poster formats and match each poster to its source paper, yielding a final dataset of $25{,}435$ verified paper--poster pairs distributed across NeurIPS (11,256), ICLR (8,504), and ICML (5,675). Each paper is further enriched with structured annotations, including a taxonomy of 15 domains, 20 tasks, and 39 methods, plus a distilled introduction summary for narrative-level retrieval (\S\ref{sec:retrieval}).

\begin{figure}[h]
\centering
\includegraphics[width=0.95\textwidth]{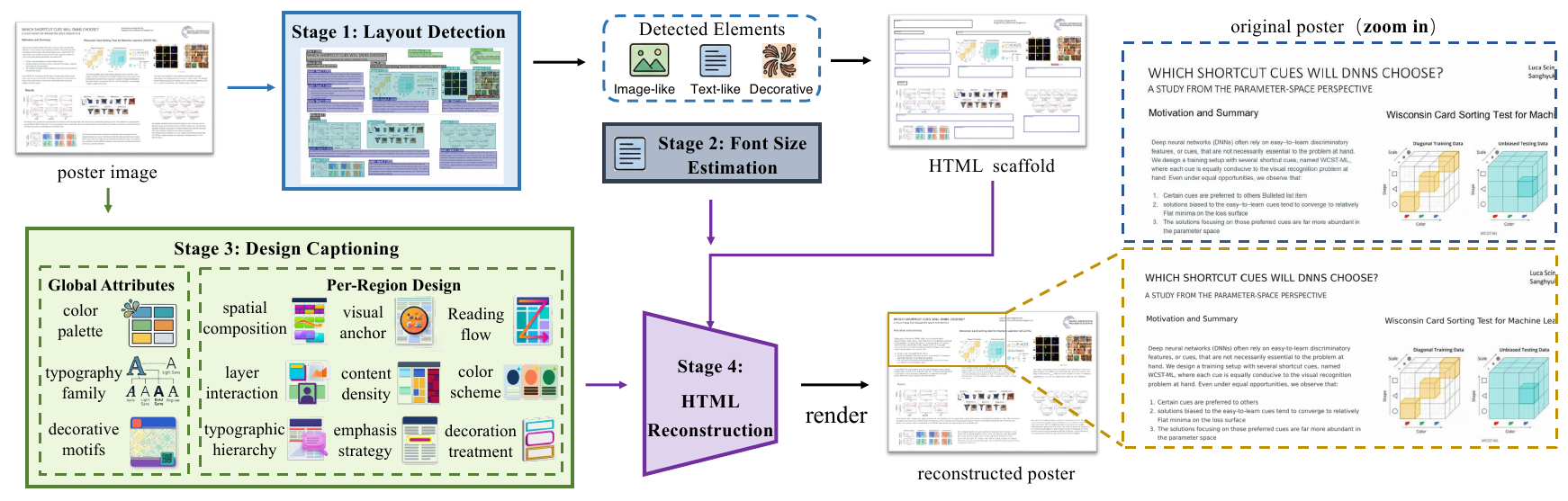}
\caption{Derendering pipeline illustration.}
\label{fig:derendering}
\end{figure}

\noindent \textbf{Derendering Pipeline}. Our derendering pipeline converts a raw
poster image into a structured HTML template with typed, fillable slots through
four stages (Figure~\ref{fig:derendering}):
\textbf{Stage 1: Layout Detection.} A fine-tuned
LayoutLMv3~\citep{huang2022layoutlmv3} model localizes all visual elements
(Title, Plain Text, Figure, Table, Formula, Caption, etc.) with bounding boxes and
confidence scores. Each detected element is subsequently assigned
to one of three semantic categories: \texttt{text-like} (titles, plain texts, captions), \texttt{image-like} (figures, tables, formulae) or
\texttt{decorative} (icons, logos).
\textbf{Stage 2: Font Size Estimation.} For each \texttt{text-like} bounding
box, the corresponding image region is cropped and processed by PaddleOCR~\citep{du2020ppocr}. Text
lines are identified by grouping OCR detection results via y-axis overlap, and
the font size in pixels is estimated as $\text{font\_size} = (h_{\text{bbox}} /
n_{\text{lines}}) \times 0.8$, where the factor $0.8$ accounts for inter-line
spacing, yielding a conservative estimate directly usable as CSS
\texttt{font-size}.
\textbf{Stage 3: Design Captioning.} The original poster image alone is passed to an MLLM, which produces a structured
design caption encoding both global attributes (the page-level color palette, typography family, and decorative motifs) and
per-region design dimensions: spatial composition, visual anchor, reading flow,
layer interaction, content density, color-scheme roles, typographic hierarchy,
emphasis strategy, and border/decoration treatment.
\textbf{Stage 4: HTML Reconstruction.} An HTML scaffold is first
programmatically constructed with absolute-positioned \texttt{<div>} elements
for each detected block: \texttt{image-like} and \texttt{decorative} regions are
cropped from the original image and embedded as \texttt{<img>} elements, while
\texttt{text-like} blocks are rendered as empty content slots with their
OCR-measured font sizes pre-annotated as styling constraints. The scaffold encodes pixel-exact block geometry but leaves all visual styling (colors, backgrounds, typography, and decorative CSS) as empty placeholders to be filled.
The scaffold, the
original poster image, and the design caption are then jointly
provided to an MLLM, which populates each \texttt{text-like} slot by
reading content directly from the poster, applies the pre-annotated font sizes and per-region descriptors, and synthesizes CSS decorations to reproduce the poster's visual appearance.

The derendering outcome is a complete HTML file in which each content block is a fully styled, addressable slot that faithfully reproduces the original poster's layout and visual appearance.

\section{Method}
\label{sec:method}

Building on our RAP dataset, we develop two applications for high-quality poster generation: \emph{PosterRAG: Training-free retrieval-augmented
generation}, which retrieves and adapts structured templates from our poster bank, and \emph{Post-training
a dedicated paper-to-poster model} with our dataset.

\begin{figure}[htbp]
\centering
\includegraphics[width=\textwidth]{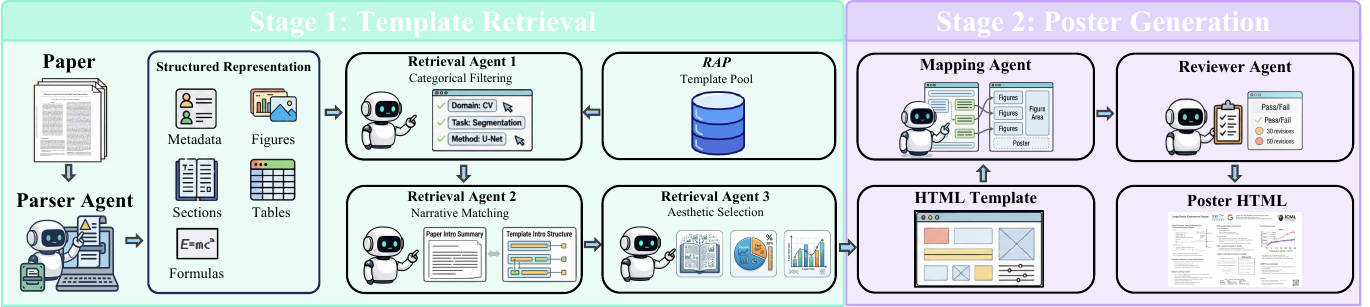}
\caption{Two-stage PosterRAG pipeline for multi-agent template retrieval and poster generation.}
\label{fig:pipeline}
\end{figure}

\subsection{PosterRAG: Training-free retrieval-augmented
generation}
\label{sec:poster_generation}

As illustrated in Figure~\ref{fig:pipeline}, PosterRAG consists of two stages: \emph{Template Retrieval} and \emph{Poster Generation}. Given an input paper $P$ in LaTeX or PDF format, Stage~1 constructs its structured representation $P_{\mathrm{meta}}$ with a \textbf{ParserAgent}  and selects the most suitable template $T^\star$ from the template bank $\mathcal{T}$ through a three-level \textbf{RetrievalAgent} cascade. Given $T^\star$ and $P_{\mathrm{meta}}$, Stage~2 produces the final poster $\hat{T}$ using a \textbf{MappingAgent} followed by an iterative \textbf{ReviewerAgent}. 

\textbf{ParserAgent.}
Following Paper2Poster~\citep{posteragent2025}, the ParserAgent takes the paper as input and converts it into a structured representation
$P_{\mathrm{meta}}=\{P_d,P_t,P_m,P_s,P_{\mathrm{parse}}\}$,
where $P_d$, $P_t$, and $P_m$ denote the domain, task, and method labels of the paper, respectively;
$P_s$ the narrative summary, and $P_{\mathrm{parse}}$  the parsed paper content, including text, figures, tables, and formulas.

\textbf{RetrievalAgent-1: Categorical filtering.}
RetrievalAgent-1 retains templates whose domain, task, and method labels match $(P_d,P_t,P_m)$. Papers with identical labels often share structural patterns, making their templates compatible with the input paper.

\textbf{RetrievalAgent-2: Narrative Matching.}
RetrievalAgent-2 re-ranks the filtered candidates by comparing the input paper's introduction summary $P_s$ with the pre-computed summary $T_s^i$ of each template's source paper. This comparison captures how papers frame their contributions and emphasize content. For example, a paper following a \emph{gap-and-fill} narrative is better matched with a template organized around the same problem--solution structure than with one based on an \emph{observation-driven} narrative. This favors templates whose organization and section emphasis align with the input paper.

\textbf{RetrievalAgent-3: Aesthetic Selection.}
RetrievalAgent-3 selects the final template by scoring each candidate poster image $I_{T^i}$ on three independent 1--5 scales: \textbf{\textit{Content Density}}, ranging from \emph{key-point} to \emph{full}; \textbf{\textit{Visualization Ratio}}, ranging from \emph{text-heavy} to \emph{figure-heavy}; and \textbf{\textit{Scientific Aesthetics}}, ranging from \emph{bare-bones} to \emph{custom-designed}. These scores form $\mathbf{s}_{T^i}\in\{1,\ldots,5\}^3$. Without user preferences, the candidate closest to the neutral midpoint is selected; otherwise, a VLM converts the preferences into dimension weights $\mathbf{w}$ to guide the selection.

\textbf{MappingAgent.}
The MappingAgent fills the typed slots $\{s_1,\ldots,s_m\}$ of the retrieved template $T^\star$ with content from $P_{\mathrm{meta}}$ in three phases. \textbf{\textit{Slot analysis}} records each slot's ID, type, label, position, and dimensions, estimating text capacity from the slot size and font size and recording the available area for images. \textbf{\textit{Content mapping and adaptation}} provides an LLM with this slot summary, the importance-ranked paper sections, figures and tables with captions, and pre-rendered assets $\mathcal{R}$. The LLM produces $\mathcal{M}=\{(s_j,\mathrm{act}_j,\mathrm{content}_j)\}$, where each action replaces, retains, or clears a slot, and the mapped content is adapted to its capacity and section type. \textbf{\textit{Programmatic filling}} then applies $\mathcal{M}$ through a deterministic dispatcher: text and image slots receive new \texttt{innerHTML} and \texttt{src} values, while unused slots are hidden. During filling, the original layout, typography, colors, and decorative styling remain unchanged, preserving the human-crafted design and full editability.

\textbf{ReviewerAgent.}
Following Paper2Poster~\citep{posteragent2025}, the ReviewerAgent takes the rendered HTML poster $\hat{T}$ as input and performs iterative visual refinement through a \emph{see-then-fix} process.

\subsection{Post-training
a dedicated paper-to-poster model}
\label{sec:post-train}

\begin{figure}[htbp]
\centering
\includegraphics[width=\textwidth]{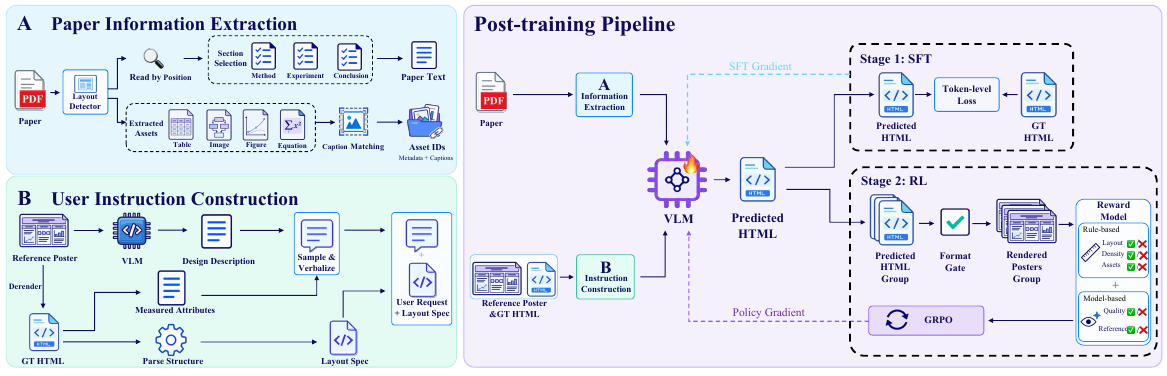}
\caption{Post-training pipeline for a vision-language model.}
\label{fig:posttrain}
\end{figure}

As illustrated in Fig.~\ref{fig:posttrain}, we construct training signals from the paper and its reference poster, including paper information and the corresponding user instruction, which serve as input to post-training. The resulting VLM enables users to generate posters tailored to their preferred design and layout.

\textbf{A. Paper Information Extraction.}
We construct the paper-side input from two complementary components: \textbf{Paper Text}  and \textbf{Paper Assets}. The paper text is extracted from method, experiments, and conclusion sections, with section-specific budgets used to retain content most relevant to poster generation. For paper assets, we detect tables, images, figures and equations from the PDF, assign unique identifiers, and associate captions based on spatial proximity. We then match the assets to visual slots in the reference poster through visual feature matching and geometric verification, using VLM adjudication for ambiguous cases and leaving uncertain matches unresolved. Assets used in the target poster are retained with their identifiers, metadata, captions, and thumbnails under a fixed visual-token budget. The resulting paper-side input is
\begin{equation}
X_{\mathrm{paper}}
=
\left(
T_{\mathrm{paper}},
\mathcal{A}_{\mathrm{paper}}
\right),
\label{eq:paper_information}
\end{equation}
where $T_{\mathrm{paper}}$ denotes the section-organized paper text and $\mathcal{A}_{\mathrm{paper}}$ denotes the selected assets and their associated information.

\textbf{B. User Instruction Construction.}
We construct the instruction-side input from two complementary components: \textbf{Design} and \textbf{Layout}. For design, the reference poster is processed by a VLM to obtain a design description covering its color palette, typography, composition, and visual emphasis. In parallel, attributes such as canvas size and visual density are measured from the ground-truth HTML. The design description and measured attributes are then sampled and verbalized with varying levels of detail and writing styles to form a natural-language user request. For layout, the ground-truth HTML is parsed to recover the columns, section order, block geometry, text--image composition, and asset assignments. This structure is converted into a layout specification without exposing the exact poster text or HTML implementation. The resulting instruction-side input is
\begin{equation}
X_{\mathrm{instruction}}
=
\left(
u_{\mathrm{design}},
s_{\mathrm{layout}}
\right),
\label{eq:instruction_information}
\end{equation}
where $u_{\mathrm{design}}$ is the natural-language user request and $s_{\mathrm{layout}}$ is the structured layout specification.

\textbf{Post-training Pipeline.}
We first perform supervised fine-tuning, followed by rendering-aware reinforcement learning with GRPO. During RL, multiple HTML candidates $\{H_i\}_{i=1}^{G}$ are sampled and rendered into poster images $I_i=\mathcal{R}(H_i)$, each receiving a composite reward:
\begin{equation}
R_i =
g(H_i)
\left[
\lambda_r R_{\mathrm{rule}}(H_i)
+
\lambda_v R_{\mathrm{vlm}}(I_i,I^\star)
\right],
\label{eq:total_reward}
\end{equation}
where $g(H_i)\in\{0,1\}$ is a validity gate that rejects invalid or non-renderable HTML, $I^\star$ is the reference poster, and $\lambda_r=0.4$ and $\lambda_v=0.6$ weight the rule-based and model-based rewards, respectively.
\textbf{Rule-based reward} $\mathrm{rule}$ uses deterministic measurements from the HTML and its rendered Document Object Model (DOM), which provides each element's position, size, and style. It contains three groups. \textbf{\textit{Layout}} checks text clipping, overlap, out-of-canvas elements, and block correspondence with the reference layout. \textbf{\textit{Density}} evaluates block and text coverage, empty regions, body font size, and the number of text blocks, preventing sparse outputs produced by deleting content or shrinking text. \textbf{\textit{Assets}} checks identifier validity, required-figure coverage, broken images, and excess empty image slots. Together, these checks enforce valid layout, sufficient content, and correct asset usage. \textbf{Model-based reward} $\mathrm{vlm}$ uses a VLM to evaluate the rendered candidate given the reference poster, paper information, and user instruction. It contains two groups. \textbf{\textit{Quality}} evaluates rendering integrity, legibility, layout, density, content completeness and faithfulness, figure usage, visual design, and instruction adherence. \textbf{\textit{Reference}} evaluates structural similarity in columns, section order, and block placement, as well as stylistic similarity in colors, typography, and decorations. The weighted dimension scores are reduced by capped penalties for visible defects, providing perceptual assessments of overall quality and reference fidelity.

\section{Experiments}
\label{sec:experiments}

\subsection{Baselines and Dataset}
We compare against Nano Banana, P2P~\citep{p2p2025}, Paper2Poster~\citep{posteragent2025}, and PosterGen~\citep{zhang2026postergen}. To ensure a fair comparison, all LLM-based agent calls across all methods are replaced with Claude-Opus4.6~\citep{anthropic2026claude}. Our evaluation uses a held-out test set of papers not included in the 25K template bank. Specifically, we construct the test set of 150 samples by collecting 10 papers from each of the 15 research domains in \DatasetName-25K.

\subsection{Evaluation Metrics}
To evaluate the diversity and human-likeness of generated posters, we assess along
two axes. 

\noindent\textbf{Diversity.}
We measure cross-paper generation diversity by computing pairwise
LPIPS~\citep{Zhang_2018_CVPR}, DINOv2-CosDist, and DISTS~\citep{ding2021dists}
over all pairs of generated posters in the test set, then averaging across pairs.
LPIPS captures perceptual diversity via AlexNet features; DINOv2-CosDist measures
semantic-level diversity using cosine distances between
DINOv2~\citep{oquab2023dinov2} embeddings; DISTS captures deep structural and
textural diversity via VGG features. Higher values indicate more varied generations.

\noindent\textbf{Human-likeness.}
We measure the distributional gap between generated posters and our
\DatasetName-25K collection using
Fréchet Inception Distance (\textbf{FID})~\citep{heusel2017gans} and
Fréchet DINOv2 Distance (\textbf{FDiD}).
FDiD replaces InceptionV3~\citep{szegedy2016inception} with
DINOv2~\citep{oquab2023dinov2} features, making it sensitive to visual
structure and layout geometry.
Together, these two metrics quantify the distribution-level fidelity of generated posters to real human-crafted designs.

\subsection{Main Results}

\begin{table}[t]
\centering
\scriptsize
\caption{Quantitative results for the two dataset-driven poster generation approaches. Best results are in \textbf{bold}, and second-best results are \underline{underlined}.}
\label{tab:generation_results}

\begin{minipage}[t]{0.49\textwidth}
\centering
\textit{(a) Template Retrieval-based Generation}
\vspace{2pt}

\renewcommand{\arraystretch}{1.12702}
\resizebox{\linewidth}{!}{%
\begin{tabular}{l ccc cc}
\toprule
\multirow{2}{*}{\textbf{Method}}
& \multicolumn{3}{c}{\textbf{Diversity}}
& \multicolumn{2}{c}{\textbf{Human-likeness}} \\
\cmidrule(lr){2-4} \cmidrule(lr){5-6}
& LPIPS$\uparrow$
& DINOv2-CosDist$\uparrow$
& DISTS$\uparrow$
& FID$\downarrow$
& FDiD$\downarrow$ \\
\midrule
NanoBanana Pro~\cite{nanobanana2025}
& 0.5695 & 0.2364 & 0.2231
& \underline{24.25} & 483.97 \\

P2P~\cite{p2p2025}
& \underline{0.5761} & \underline{0.2847} & 0.2340
& 33.24 & \underline{377.25} \\

Paper2Poster~\cite{posteragent2025}
& 0.5547 & 0.1138 & 0.2107
& 81.59 & 1156.73 \\

PosterGen~\cite{zhang2026postergen}
& 0.4131 & 0.2710 & \underline{0.2527}
& 133.83 & 682.27 \\
\midrule

\rowcolor{azure}
Ours (Image Guidance)
& 0.5584 & 0.0877 & 0.2246
& 77.11 & 1267.84 \\

\rowcolor{lavender}
Ours (HTML Guidance)
& \textbf{0.6471}
& \textbf{0.4443}
& \textbf{0.3490}
& \textbf{23.75}
& \textbf{306.79} \\
\bottomrule
\end{tabular}%
}
\end{minipage}
\hfill
\begin{minipage}[t]{0.49\textwidth}
\centering
\textit{(b) Post-trained Paper-to-Poster Models}
\vspace{2pt}

\resizebox{\linewidth}{!}{%
\begin{tabular}{l ccc cc}
\toprule
\multirow{2}{*}{\textbf{Model}}
& \multicolumn{3}{c}{\textbf{Diversity}}
& \multicolumn{2}{c}{\textbf{Human-likeness}} \\
\cmidrule(lr){2-4} \cmidrule(lr){5-6}
& LPIPS$\uparrow$
& DINOv2-CosDist$\uparrow$
& DISTS$\uparrow$
& FID$\downarrow$
& FDiD$\downarrow$ \\
\midrule
GPT-6-Astra~\cite{openai_gpt6_astra}
& 0.5704 & 0.2093 & 0.2369
& 27.64 & 341.64 \\

Kimi-K3~\cite{kimiteam2026kimik3openfrontier}
& 0.5856 & 0.1974 & 0.2445
& 27.94 & 340.50 \\

Gemini-3.1-Pro~\cite{google2026gemini31}
& 0.5956 & 0.1983 & 0.2365
& 41.49 & 408.05 \\

Claude-Opus-5~\cite{anthropic2026claudeopus5}
& 0.6414 & 0.2320 & 0.2728
& \textbf{21.56} & \textbf{289.49} \\
\midrule

Qwen3.8-27B~\cite{qwen38}
& 0.6284 & 0.2579 & 0.2680
& 32.44 & 465.00 \\

\rowcolor{azure}
Qwen3.8-27B-SFT (Ours)
& \underline{0.6599}
& \textbf{0.2784}
& \underline{0.2892}
& 28.00
& 326.24 \\

\rowcolor{lavender}
Qwen3.8-27B-RL (Ours)
& \textbf{0.6925}
& \underline{0.2735}
& \textbf{0.2944}
& \underline{23.42}
& \underline{306.94} \\
\bottomrule
\end{tabular}%
}
\end{minipage}
\end{table}

\subsubsection{Quantitative Comparison}

Table~\ref{tab:generation_results}(a) shows that RAP (HTML Guidance) achieves the best scores across all five metrics.
It attains FID/FDiD of 23.75/306.79, with FDiD 18.7\% lower than P2P (377.25), the best baseline on this metric.
Its DINOv2-CosDist reaches 0.4443, compared with 0.2847 for P2P and 0.1138 for Paper2Poster.
These results support retrieval and adaptation of human-crafted templates for improving both cross-paper diversity and distributional human-likeness.

Table~\ref{tab:generation_results}(b) shows that RAP-based SFT improves all five metrics over the base Qwen3.8-27B.
Subsequent RL reduces FID/FDiD from 28.00/326.24 to 23.42/306.94 and increases LPIPS and DISTS, with a small decrease in DINOv2-CosDist.
Among the compared models, our RL variant achieves the highest LPIPS and DISTS and the second-lowest FID/FDiD, behind Claude-Opus-5.
These results demonstrate the utility of RAP for improving cross-paper diversity and distributional alignment with human-crafted posters through post-training.

\subsubsection{Qualitative Comparison}

\begin{figure}[htbp]
  \centering
  \includegraphics[width=\textwidth]{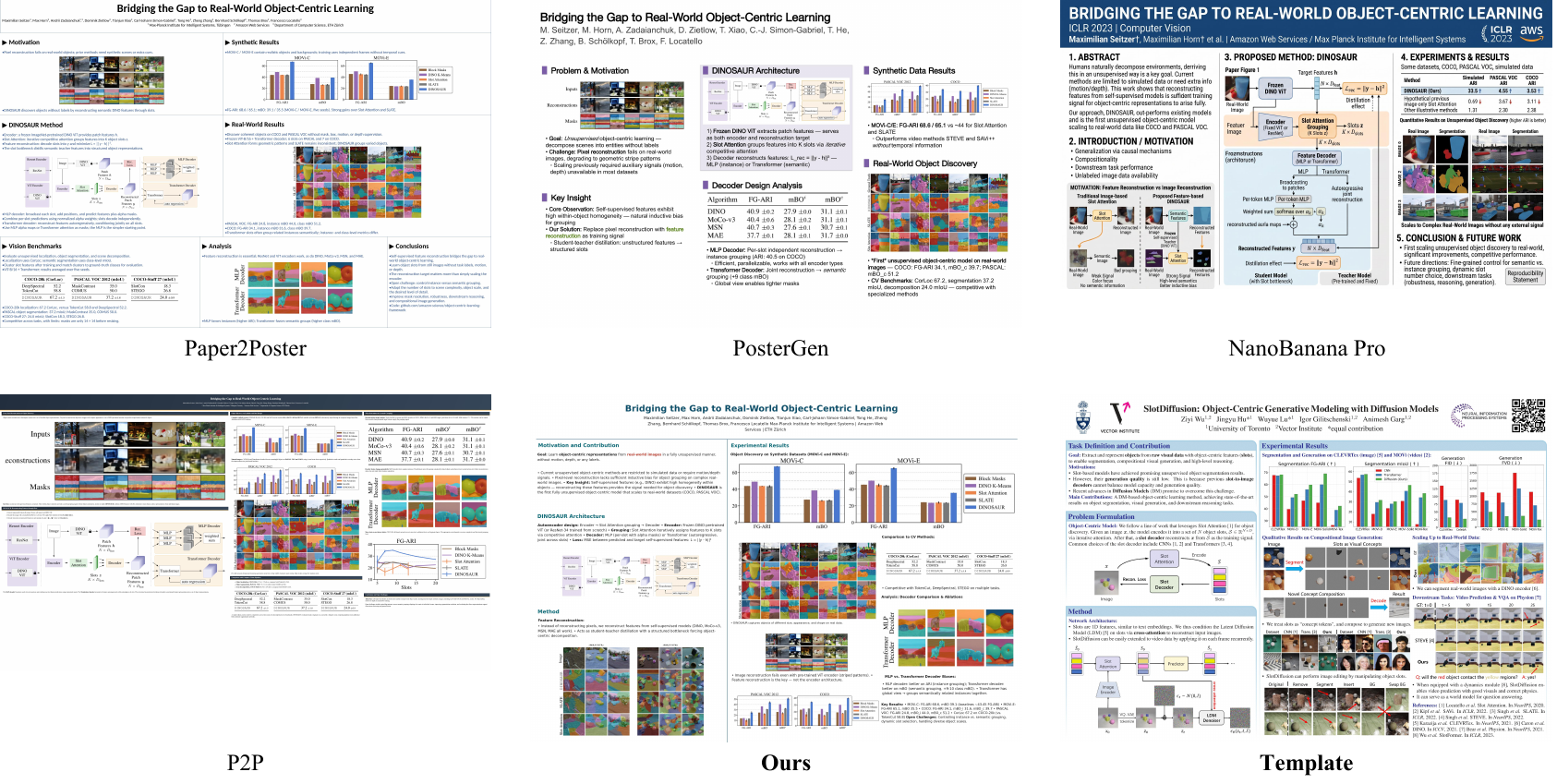}
  \caption{Qualitative comparison between PosterRAG (ours) and baseline methods.}
  \label{fig:qualitative_rag}
\end{figure}

\begin{figure}[htbp]
  \centering
  \includegraphics[width=\textwidth]{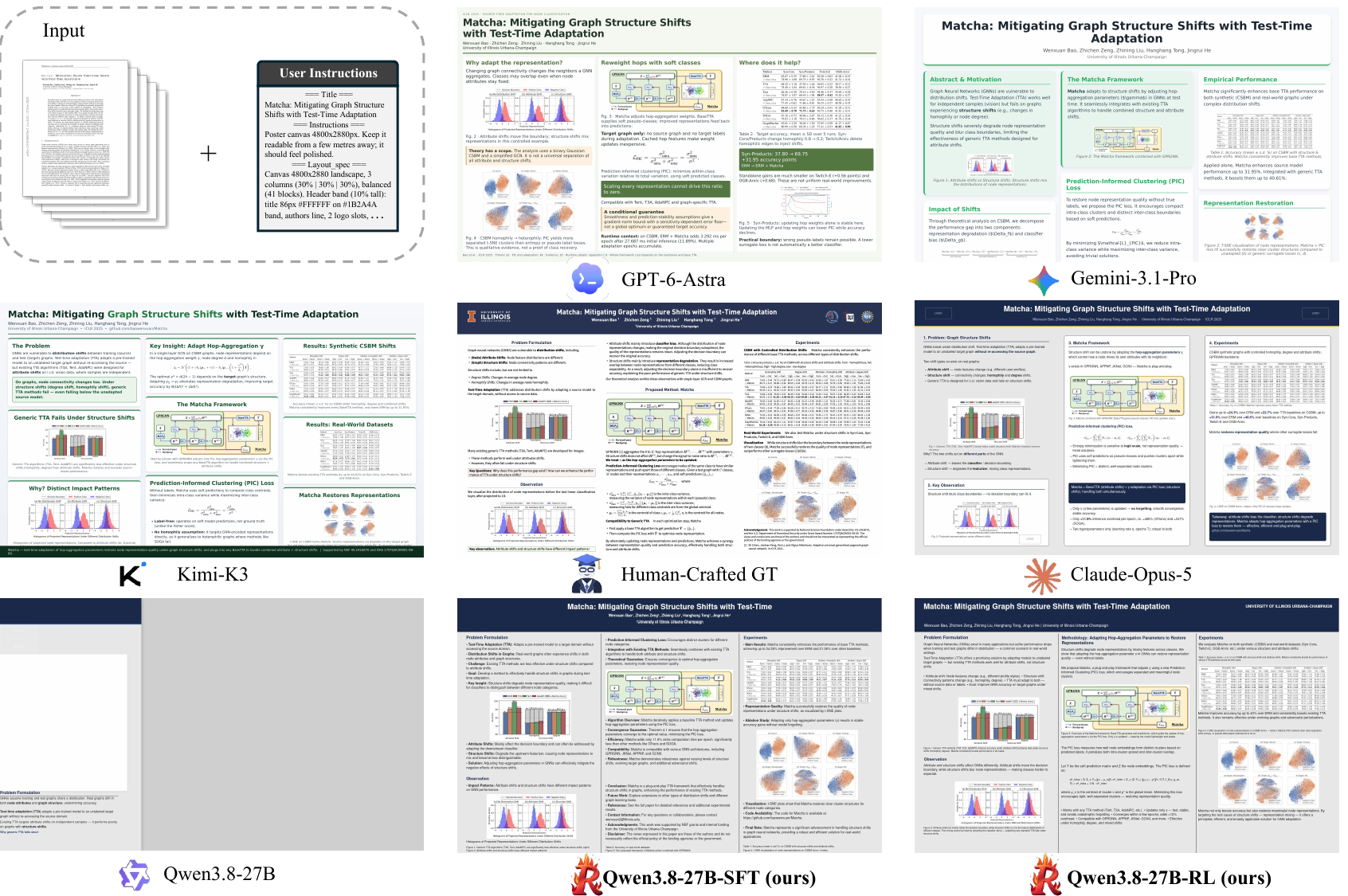}
  \caption{Qualitative comparison between our post-trained model and other vision-language models.}
  \label{fig:qualitative_post}
\end{figure}

Figure~\ref{fig:qualitative_rag} compares posters generated from the same paper.
Paper2Poster uses a conventional column-based layout, whereas NanoBanana Pro produces a more stylized raster poster.
RAP (HTML Guidance) follows the retrieved template's color palette, typographic hierarchy, and section organization while adapting its content to the input paper.
This example illustrates how human-crafted templates provide reusable design guidance for paper-to-poster generation.

Figure~\ref{fig:qualitative_post} compares outputs for the same paper and user instruction.
The base Qwen3.8-27B produces a largely incomplete poster, whereas both post-trained variants organize text, figures, and tables into complete three-column layouts.
Their dark title bands and column organization follow the supplied design specification and resemble the human-crafted reference.
This example illustrates improved layout adherence and more complete rendered outputs after post-training.

\subsection{Ablation Studies}

RL reward-component ablations are reported in Appendix~\ref{app:rl_reward_ablation}.

\noindent\textbf{Ablation on HTML Generation.}
We compare \textit{Ours (HTML Guidance)}, which fills retrieved HTML templates, with \textit{Ours (Image Guidance)}, which generates HTML from poster-image references.
HTML Guidance improves all five metrics, reducing FID/FDiD from 77.11/1267.84 to 23.75/306.79.
DINOv2-CosDist rises from 0.0877 to 0.4443, alongside higher LPIPS (0.5584 to 0.6471) and DISTS (0.2246 to 0.3490).
Unlike image-only guidance, the HTML scaffold explicitly constrains layout and styling while allowing content replacement and refinement.
These results support using reconstructed HTML, rather than the reference image alone, to improve distributional human-likeness and cross-paper diversity.

\noindent\textbf{Ablation on derendering.}
For this poster reconstruction task, we adopt six metrics:
\textbf{SSIM}~\cite{wang2004ssim} measures pixel-level structural similarity.
\textbf{DISTS}~\cite{ding2021dists} measures deep structural and texture
dissimilarity.
\textbf{Layout mIoU} computes mean IoU between detected and ground-truth layout
elements, in both class-aware (mIoU) and class-agnostic (mIoU$_\text{ag}$)
variants.
\textbf{FID}~\citep{heusel2017gans} and \textbf{FDiD} measure the distributional fidelity to the original poster set.

\begin{table}[htbp]
\centering
\footnotesize
\caption{Ablation studies on the derendering pipeline. Best results in \textbf{bold}.}
\label{tab:ablation}
\begin{minipage}[t]{0.4325\textwidth}
\centering
\textit{(a) Backbone Model}
\resizebox{\textwidth}{!}{%
\begin{tabular}{l cccc cc}
\toprule
& \multicolumn{4}{c}{Pairwise} & \multicolumn{2}{c}{Fidelity} \\
\cmidrule(lr){2-5} \cmidrule(lr){6-7}
& SSIM$\uparrow$ & DISTS$\downarrow$ & mIoU$\uparrow$ & mIoU$_\text{ag}$$\uparrow$ & FID$\downarrow$ & FDiD$\downarrow$ \\
\midrule
GPT-4o          & 0.7928 & 0.1643 & 0.4692 & 0.5081 & 41.00  & 371.41 \\
Qwen-3.5        & 0.7922 & 0.0966 & 0.4349 & 0.4702 & 28.86  & 120.18 \\
Gemini-3.1-Pro  & \textbf{0.8076} & \textbf{0.0623} & 0.6400 & 0.6706 & 17.30  & 66.28 \\
\rowcolor{gray!20}
\cellcolor{lavender}Claude-Opus-4.6 & \cellcolor{lavender}0.7896 & \cellcolor{lavender}0.0663 & \cellcolor{lavender}\textbf{0.6897} & \cellcolor{lavender}\textbf{0.7246} & \cellcolor{lavender}\textbf{16.90} & \cellcolor{lavender}\textbf{61.71} \\
\bottomrule
\end{tabular}%
}
\end{minipage}
\hfill
\begin{minipage}[t]{0.5275\textwidth}
\centering
\textit{(b) Pipeline Components (backbone: Claude-Opus-4.6)}
\vspace{2pt}
\resizebox{\textwidth}{!}{%
\begin{tabular}{l cccc cc}
\toprule
& \multicolumn{4}{c}{Pairwise} & \multicolumn{2}{c}{Fidelity} \\
\cmidrule(lr){2-5} \cmidrule(lr){6-7}
& SSIM$\uparrow$ & DISTS$\downarrow$ & mIoU$\uparrow$ & mIoU$_\text{ag}$$\uparrow$ & FID$\downarrow$ & FDiD$\downarrow$ \\
\midrule
Base                              & 0.5843 & 0.4265 & 0.0652 & 0.1069 & 53.63 & 977.20 \\
\quad +Layout                     & 0.7744 & 0.0857 & 0.5236 & 0.5590 & 18.32 & 93.48  \\
\quad +Layout +OCR                & 0.7859 & 0.0686 & 0.6808 & 0.7148 & 17.42 & 65.49  \\
\rowcolor{lavender}
\quad +Layout +OCR +Design (Ours) & \textbf{0.7896} & \textbf{0.0663} & \textbf{0.6897} & \textbf{0.7246} & \textbf{16.90} & \textbf{61.71} \\
\bottomrule
\end{tabular}%
}
\end{minipage}
\end{table}

\noindent\textit{Backbone model.}
Table~\ref{tab:ablation}(a) compares four backbone MLLMs under the full pipeline.
Claude-Opus-4.6~\citep{anthropic2026claude} and
Gemini-3.1-Pro~\citep{google2026gemini31} clearly outperform
GPT-4o~\citep{openai2024gpt4o} and Qwen-3.5~\citep{qwen35}.
Gemini-3.1-Pro achieves the best reconstruction fidelity under SSIM and DISTS, while
Claude-Opus-4.6 leads on layout accuracy and distributional fidelity,
suggesting it better preserves the structural organization of the original
poster. We therefore adopt Claude-Opus-4.6 as the backbone for all other
experiments.

\noindent\textit{Pipeline components.}
Table~\ref{tab:ablation}(b) shows that adding Layout yields the largest improvement across all metrics, raising mIoU from 0.0652 to 0.5236.
OCR further improves layout accuracy and distributional fidelity, while Design Caption provides smaller but consistent gains across all six metrics.
The full pipeline achieves the best reconstruction scores, supporting the complementary roles of geometric, typographic, and design constraints.

\FloatBarrier
\section{Conclusion}
\label{sec:conclusion}

We have cast paper-to-poster generation as a visual coding problem and constructed the resource it requires. \DatasetNameFull{} contains 25{,}435 verified paper-poster pairs from NeurIPS, ICML, and ICLR (2022--2026), annotated along a three-level taxonomy of 15 domains, 20 tasks, and 39 methods, together with a held-out benchmark of 150 papers whose human posters are withheld for evaluation. A fully automated derendering pipeline rewrites each poster as styled HTML/CSS with typed, refillable slots, reconstructing it faithfully enough (SSIM~$0.79$, layout mIoU~$0.69$) that fidelity can be measured by rendering and comparing rather than by judge scoring.

These paper-poster-markup triplets support two applications. \MethodName{} is training-free: it retrieves a human-designed template and refills its slots without disturbing the layout or styling, and it leads every baseline on both diversity and human-likeness. An SFT+RL recipe trains a dedicated model against the same references. Together they show that human designs are not merely evaluation yardsticks but a reusable source of design knowledge. We release the dataset, derendering pipeline, codebase, and an online demo to make reference-grounded visual coding a reproducible setting for future research.

\subsection*{AI use statement}
Claude-Opus-4.6 powered our derendering and RAG pipelines and the agentic baselines; derendering ablations additionally evaluated GPT-4o, Qwen-3.5, and Gemini-3.1-Pro. NanoBanana Pro served as an image-generation baseline. We post-trained Qwen3.8-27B through SFT and RL, using Gemini-3.8-Flash for VLM-based rewards, and compared against its base model, GPT-6-Astra, Kimi-K3, Gemini-3.1-Pro, and Claude-Opus-5. VLMs also assisted training-data construction. AI assistants, including Codex, supported manuscript drafting and language polishing. We retain full responsibility for all research decisions and submitted content.

\clearpage
{
\small
\bibliographystyle{unsrtnat}
\bibliography{references}
}

\clearpage
\appendix
\section{Additional Dataset Statistics}
\label{app:dataset}

\subsection{Dataset Overview and Taxonomy Distribution}
\label{app:taxonomy}

\DatasetName-25K annotates each paper with a three-level taxonomy comprising 15 domains, 20 tasks, and 39 methods. Table~\ref{tab:taxonomy_domains} lists all domain categories with their counts and proportions.

Alongside the domain statistics, Figure~\ref{fig:rap_dataset_overview} provides a visual overview of human-crafted posters in RAP. Their varied layouts and visual styles form the reference pool used by our retrieval and post-training applications.

\noindent\begin{minipage}{\textwidth}
\captionsetup{hypcap=false}
\centering
\includegraphics[width=\textwidth]{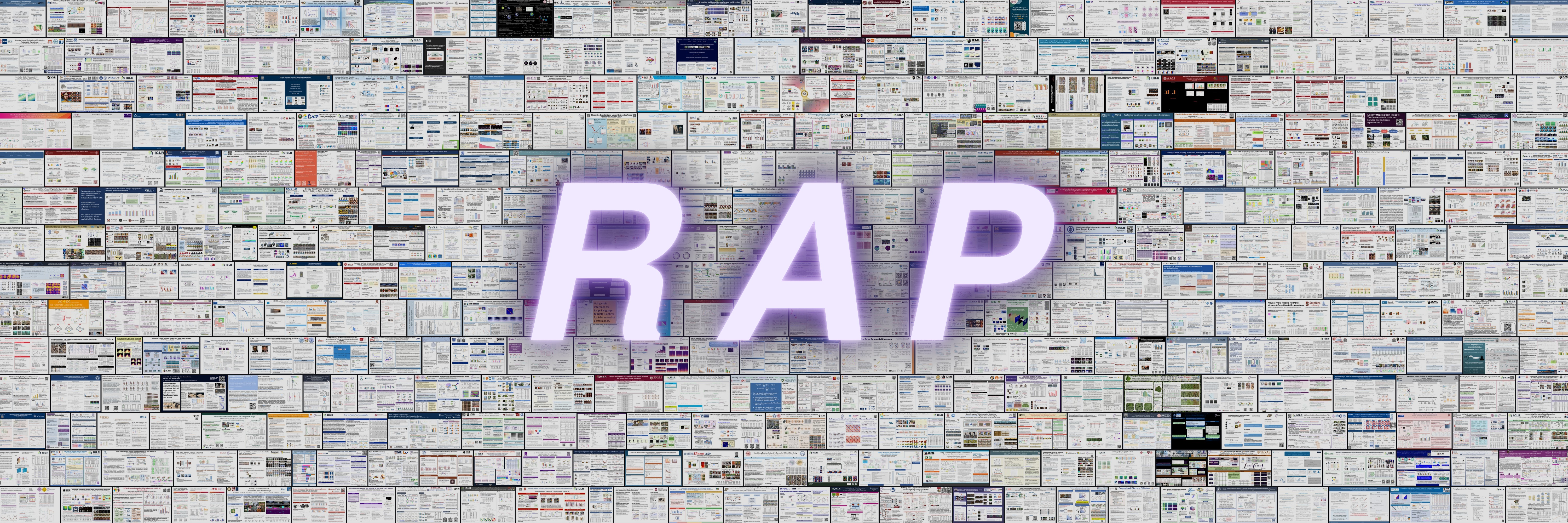}
\captionof{figure}{Overview of human-crafted posters in RAP. The mosaic illustrates variation in layout, color palette, typography, and content density.}
\label{fig:rap_dataset_overview}
\end{minipage}

\begin{table}[h]
\centering
\caption{Full domain taxonomy of \DatasetName-25K (15 categories).}
\label{tab:taxonomy_domains}
\small
\begin{tabular}{lrr}
\toprule
\textbf{Domain} & \textbf{Count} & \textbf{\%} \\
\midrule
Optimization \& Learning Theory & 3,487 & 13.7 \\
AI Safety \& Alignment & 2,580 & 10.1 \\
Computer Vision & 2,414 & 9.5 \\
Reinforcement Learning & 2,312 & 9.1 \\
Multimodal \& Cross-modal Learning & 2,164 & 8.5 \\
NLP \& LLMs & 1,966 & 7.7 \\
Representation \& Transfer Learning & 1,870 & 7.4 \\
AI for Natural Sciences & 1,672 & 6.6 \\
Probabilistic \& Generative ML & 1,665 & 6.6 \\
Graph \& Relational Learning & 992 & 3.9 \\
Data-centric ML \& Evaluation & 926 & 3.6 \\
ML Systems \& Infrastructure & 899 & 3.5 \\
AI for Health \& Social Sciences & 894 & 3.5 \\
Robotics \& Embodied AI & 872 & 3.4 \\
Fairness \& Social Impact & 722 & 2.8 \\
\midrule
\textbf{Total} & \textbf{25,435} & \textbf{100.0} \\
\bottomrule
\end{tabular}
\end{table}

\noindent\begin{minipage}{\textwidth}
\subsection{Per-Conference Breakdown}
\label{app:per_conf}

Table~\ref{tab:per_conf} provides the detailed per-conference and per-year breakdown for \DatasetName-25K. Note that for 2026, only ICLR data is currently included.

\begin{table}[H]
\centering
\caption{Per-conference and per-year poster counts.}
\label{tab:per_conf}
\small
\begin{tabular}{lrrrrr|r}
\toprule
\textbf{Conference} & \textbf{2022} & \textbf{2023} & \textbf{2024} & \textbf{2025} & \textbf{2026} & \textbf{Total} \\
\midrule
NeurIPS & 2,280 & 2,613 & 3,037 & 3,326 & 0 & 11,256 \\
ICLR    & 830   & 877   & 1,447 & 2,085 & 3,265 & 8,504 \\
ICML    & 882   & 1,471 & 1,751 & 1,571 & 0 & 5,675 \\
\midrule
\textbf{Total} & \textbf{3,992} & \textbf{4,961} & \textbf{6,235} & \textbf{6,982} & \textbf{3,265} & \textbf{25,435} \\
\bottomrule
\end{tabular}
\end{table}
\end{minipage}

\section{Implementation Details}
\label{app:implementation}

\subsection{Method}
\label{app:method}

\subsubsection{POSTERRAG: TRAINING-FREE RETRIEVAL-AUGMENTED GENERATION}
\label{app:method_rag}

\paragraph{Stage 1: Template Retrieval.}
\label{sec:retrieval}

Stage~1 selects the most suitable poster template $T^*$ from the template bank $\mathcal{T}$ for a given input paper.

\textbf{ParserAgent.}
The ParserAgent converts the raw paper into a structured representation $P_{\text{meta}}$ that drives all downstream retrieval and generation.
The parser supports both LaTeX source and PDF input, extracting metadata, segmenting content by section boundaries, and building a figure registry $\mathcal{F}$ of all image assets in PNG format for HTML embedding; the result is a parsed content object $P_{\text{parse}} = \{\text{meta}, \text{abstract}, \text{sections}, \text{figures}, \text{tables}, \text{formulas}\}$.
Beyond content extraction, the ParserAgent classifies the paper into a three-level taxonomy by prompting an LLM with the abstract and section titles, producing a \emph{domain} label $P_d$, a \emph{task} label $P_t$, and a \emph{method} label $P_m$.
It also distills the introduction into a structured summary $P_{\text{s}}$ along four dimensions: motivation, problem formulation, key insight, and claimed contributions, which serves as the semantic fingerprint for narrative-level template matching.
Together these form $P_{\text{meta}} = \{P_d,\, P_t,\, P_m,\, P_{s},\, P_{\text{parse}}\}$.

\textbf{RetrievalAgent-1: Categorical filtering.}
RetrievalAgent-1 performs a hard filter, retaining only those templates whose categorical labels exactly match the input paper's domain, task, and method.
The rationale is that papers sharing the same domain, task, and method tend to follow similar paper structures and argue their contributions in comparable ways, making their posters naturally compatible in layout and content organization.

\textbf{RetrievalAgent-2: Narrative Matching.}
Beyond category-level structural compatibility, this stage performs finer-grained selection by comparing how papers frame their contributions and what content they emphasize.
Each template's source paper is associated with a pre-computed introduction summary $T^{i}_s$ extracted during dataset construction;
RetrievalAgent-2 measures the narrative similarity between the input paper's
$P_s$ and each candidate's $T^{i}_s$, and re-ranks the candidates accordingly.
This captures whether two papers share similar narrative logic in their introductions.
For instance, a paper following a \textit{gap-and-fill} structure
(``existing methods fail to handle X; we propose Y to address this'')
pairs better with a template whose source paper argues along the same
problem--solution arc, rather than one structured as an
\textit{observation-driven} narrative
(``experiments reveal that phenomenon X occurs under condition Y;
motivated by this finding, we derive Z'').
By matching at the narrative level, RetrievalAgent-2 ensures that the selected
template's content layout and section emphasis align with how the input paper
tells its story.

\textbf{RetrievalAgent-3: Aesthetic selection.}
The final retrieval stage selects a single template by assessing visual style rather than content.
RetrievalAgent-3 scores the poster image $I_{T^i}$ of each candidate on three mutually independent dimensions on a 1--5 integer scale, producing a score vector $\mathbf{s}_{T^i} \in \{1,\dots,5\}^3$:
\textbf{Content Density}, from \textit{key-point} to \textit{full}, capturing how much of the paper's content is reproduced;
\textbf{Visualization Ratio}, from \textit{text-heavy} to \textit{figure-heavy}, capturing the balance between figures/tables and textual content; and
\textbf{Scientific Aesthetics}, from \textit{bare-bones} to \textit{custom-designed}, capturing the level of decoration and visual polish.
Without user preference, the candidate closest to the neutral midpoint is selected as a safe default.
When style preferences are provided, a VLM-based intent recognizer converts them into per-dimension weights $\mathbf{w}$, shifting selection toward user-emphasized dimensions.

\paragraph{Stage 2: Poster Generation.}
\label{sec:generation}

Given the retrieved template $T^*$ and the parsed paper content $P_{\text{meta}}$, Stage~2 produces the final poster via a \textbf{MappingAgent} that fills the template with paper content, followed by an iterative \textbf{ReviewerAgent} that refines visual quality.

\textbf{MappingAgent.}
The MappingAgent takes the retrieved template $T^*$ with its typed slots $\{s_1, s_2, \dots, s_m\}$ and the parsed paper $P_{\text{meta}}$, and produces a filled poster through three consecutive phases.
\textbf{\textit{Slot analysis.}}
The agent profiles each template slot programmatically.
For text-like slots, it computes an approximate character capacity derived from their display dimensions $(w_j, h_j)$ and font size $f_j$; for image-like slots, it records the available dimensions for figure sizing.
The result is a slot summary describing each slot's ID, type, label, spatial position, dimensions, and text capacity.
\textbf{\textit{Content mapping and adaptation.}}
An LLM receives the slot summary alongside the paper's section content with importance scores, figure and table list with captions, and any pre-rendered equation or table assets $\mathcal{R}$.
It produces a mapping $\mathcal{M} = \{(s_j, \text{act}_j, \text{content}_j)\}$ for every slot, where $\text{act}_j \in \{\texttt{replace\_text}, \texttt{replace\_image}, \texttt{replace\_rendered}, \texttt{keep}, \texttt{clear}\}$ is the fill action, and $\text{content}_j$ is adapted from the corresponding source paper element to fit the slot's capacity and formatted according to section type (e.g., bullet points for methods, flowing prose for introductions).
\textbf{\textit{Programmatic filling.}}
The core insight of our approach is that filling is performed \emph{programmatically} via direct manipulation rather than LLM-based HTML regeneration.
Using a deterministic action dispatcher, only slot-level content is updated: text slots receive new \texttt{innerHTML}, image slots receive new \texttt{src} attributes from the figure registry $\mathcal{F}$, and unmapped or stale slots are hidden via visibility flags.
All CSS rules, spatial layout, typography, color schemes, gradients, and decorative elements remain exactly as in the original human-created template.
This structure-preserving invariant ensures that RAP inherits professional visual design quality while maintaining full editability.

\textbf{ReviewerAgent.}
Although programmatic filling preserves the template's global structure, inserting new content into fixed slots may still cause visual artifacts such as text overflow, font-size mismatches, poor color contrast, or awkward whitespace.
The ReviewerAgent resolves these issues through an iterative \emph{see-then-fix} loop where each round renders the current HTML poster $\hat{T}$ into a screenshot and uses VLM feedback to guide localized refinements.
\textbf{\textit{Visual review.}}
At each round $r$, a headless Chromium browser renders $\hat{T}$ to a pixel-accurate screenshot $I_r$.
The VLM receives $I_r$ along with past fixes, returning an overall score $\text{score}_r$ and a list of block-level repair suggestions $\{(b_j, \text{issue}_j, \text{patch}_j)\}$ specifying the target block ID, issue type, and a targeted CSS patch.
\textbf{\textit{Constrained refinement.}}
Each $\text{patch}_j$ is applied locally to its target block $b_j$ in $\hat{T}$ under strict safety constraints that explicitly block destructive modifications across four categories: visibility-hiding rules that would render content invisible to users, text-hostile overflow and truncation properties that cause text to be clipped or hidden, layout geometry alterations that would shift or resize elements, and structural changes to the DOM or color scheme that exceed the intended scope of local refinement.
\textbf{\textit{Termination and design rationale.}}
The loop terminates once $\text{score}_r \geq \tau_{\text{score}}$ (default: 8/10), $N_{\max}=3$ rounds are reached, or no further patches are proposed by the agent.
Two decisions distinguish this agent from the prior methods~\citep{posteragent2025, p2p2025}: \emph{programmatic CSS patching} rather than full HTML regeneration ensures layout stability, and \emph{screenshot-level feedback} detects rendering-dependent defects invisible from source code alone.

\subsection{Prompt Templates}
\label{app:prompts}

\subsubsection{Prompt 1: Design Caption in Derendering}
\label{app:prompt_design_caption}

Used in Stage~3 of the derendering pipeline to extract a structured design
intermediate representation (IR) from a poster image, \emph{without} any
pre-computed bounding boxes.

\paragraph{System prompt.}
\emph{``You are a senior visual designer and information architect
specialised in academic posters.
Return a strict JSON description that downstream code can render
back to HTML faithfully.''}

\paragraph{User prompt.}
The MLLM receives a single input: the original poster image.
It is asked to identify high-level semantic regions entirely from visual
observation (no layout annotations are provided) and, for each region,
assign design tokens drawn from a fixed vocabulary of
\textbf{10 design dimensions} (9 per-region + 1 global):

\begin{itemize}[leftmargin=*,noitemsep]
  \item \texttt{global\_decorations}: \textit{solid\_block\_carrier /
        thin\_divider\_lines / logo\_placeholder /
        qr\_placeholder / background\_pattern}
  \item \texttt{composition}: \textit{grid\_2col / grid\_3col /
        flex\_row / flex\_col / absolute\_overlap}
  \item \texttt{visual\_anchor}: \textit{scale\_dominant /
        weight\_dominant / contrast\_dominant / isolated\_focal}
  \item \texttt{flow}: \textit{z\_pattern / f\_pattern /
        top\_down / left\_right}
  \item \texttt{interaction}: \textit{solid\_carrier /
        offset\_underlay / partial\_overlay /
        glass\_morphism / flat\_inline}
  \item \texttt{density}: \textit{airy / balanced / dense}
  \item \texttt{color\_scheme}: \textit{neutral\_palette /
        warm\_palette / cool\_palette}
  \item \texttt{typography}: \textit{sans\_modern /
        serif\_classic / mono\_technical}
  \item \texttt{emphasis}: \textit{bold\_caps / color\_tint /
        icon\_prefix / highlight\_bg}
  \item \texttt{decoration}: \textit{border\_solid\_square /
        border\_dashed\_square / border\_solid\_rounded /
        border\_none\_shadow / border\_thick\_accent}
\end{itemize}

\paragraph{Key rules.}
\begin{itemize}[leftmargin=*,noitemsep]
  \item \textbf{Region granularity}: use \textbf{3--7 regions}, one per
        major section (Header / Motivation / Method / Results / Conclusion).
        Sub-figures, columns, and sub-panels within the same section
        must \emph{not} be split into separate regions.
  \item \textbf{Dimension notes}: for every dimension entry, a concrete
        \texttt{notes} sentence is required describing the actual visual
        appearance (specific colours, paddings, weight hierarchy);
        restating the variant name verbatim is prohibited.
  \item \textbf{Colour palette}: the global palette must name four roles---
        \texttt{surface} (background), \texttt{primary} (brand colour),
        \texttt{accent} (highlight), \texttt{fg} (foreground text)---each
        as a hex value observed in the image.
  \item \textbf{Design caption}: each region must include a \texttt{design\_caption}
        of 3--5 sentences describing its visual design holistically
        (colour usage, typographic hierarchy, density, decoration),
        without restating per-dimension variant names.
\end{itemize}

\paragraph{Output.}
A single JSON object with fields \texttt{global\_caption},
\texttt{global} (palette, typography, decorations), and a
\texttt{regions} array.
Each element of \texttt{regions} contains the region type,
title, reading order, content caption, all 9 per-region dimension tokens
(each a \texttt{\{value, notes\}} pair), and a free-form
\texttt{design\_caption}.
No markdown fences, no commentary, no think blocks.

\subsubsection{Prompt 2: HTML Reconstruction in Derendering}
\label{app:prompt_html_gen}

Used in Stage~4 of the derendering pipeline to convert the layout
and design IR into a pixel-faithful HTML/CSS reproduction of the poster.

\paragraph{System prompt.}
\emph{``You are an HTML/CSS reconstruction expert.
Your goal: make the rendered HTML look \textbf{identical} to the
attached reference image.''}

\paragraph{User prompt.}
The MLLM receives five inputs:
(1)~the \textbf{reference poster image} (visual ground truth);
(2)~a \textbf{layout JSON} containing, for every detected region,
its bounding-box coordinates (xyxy, in pixels), semantic class label
(title / plain text / figure / table / formula / \ldots),
OCR-extracted text content, and OCR-measured font size
(computed as $\text{crop\_height}/n_{\text{lines}}\times 0.8$);
(3)~the \textbf{design specification} produced by Prompt~1,
comprising the global colour palette, typography family,
page-level decoration types, and per-region design tokens
(\texttt{composition}, \texttt{visual\_anchor}, \texttt{flow},
\texttt{interaction}, \texttt{density}, \texttt{color\_scheme},
\texttt{typography}, \texttt{emphasis}, \texttt{decoration})
each paired with a free-form \texttt{notes} sentence;
(4)~\textbf{dimension$\to$CSS hints}: for each enum variant the recommended
CSS pattern (e.g.\ \texttt{solid\_carrier}$\Rightarrow$a
\texttt{position:absolute} background \texttt{<div>} at \texttt{z-index:0}
with a coloured title strip; scaffold content divs raised to
\texttt{z-index:2});
(5)~\textbf{font-size hints}: per-block target \texttt{font-size} values;
entries tagged ``[measured by OCR]'' are exact targets and must be
reproduced precisely in CSS.

\paragraph{Key rules.}
\begin{itemize}[leftmargin=*,noitemsep]
  \item \textbf{Geometry lock}: every block's
        \texttt{left}/\texttt{top}/\texttt{width}/\texttt{height}
        is immutable; the MLLM must not add, remove, or reorder scaffold
        \texttt{<div>} elements, and must not alter canvas dimensions.
  \item \textbf{Text fidelity}: for each text-like block, the MLLM must
        read the \emph{exact text} from the reference image---reproducing
        every word, symbol, capitalisation, and line break.
        Mathematical expressions must be preserved using Unicode symbols,
        \texttt{<sup>}/\texttt{<sub>}, and CSS alignment; text must never
        overflow its bounding box.
  \item \textbf{Style matching}: the MLLM must match font family
        (sans-serif / serif / monospace), font size (using the OCR-measured
        value as target), font weight and style, text colour, alignment,
        and line spacing as observed in the reference image.
  \item \textbf{Region backgrounds}: when a region's \texttt{interaction}
        token is not \texttt{flat\_inline} or its \texttt{decoration} token
        is not \texttt{border\_none\_shadow}, extra background
        \texttt{<div>} elements must be injected inside the canvas
        (z-index~0 for the outer coloured frame; z-index~1 for an inner
        white box if noted by the design caption) enclosing that region's
        blocks; scaffold block divs are then raised to z-index~2.
  \item \textbf{Visual reproduction}: all backgrounds, gradients, borders,
        divider lines, decorative shapes, header bars, and shadows must be
        reproduced using only CSS (\texttt{background}, \texttt{border},
        \texttt{box-shadow}, \texttt{border-radius},
        \texttt{linear-gradient}, \texttt{::before}/\texttt{::after});
        no external images or scripts are permitted.
  \item \textbf{Design spec precedence}: the design specification from
        Prompt~1 takes precedence over the MLLM's own interpretation of
        the reference image whenever they conflict.
\end{itemize}

\paragraph{Output.}
Complete HTML from \texttt{<!doctype html>} to \texttt{</html>}.
No markdown fences, no explanations, no \texttt{<script>} tags,
no external URLs.

\subsubsection{Prompt 3: Paper Parser (ParserAgent)}
\label{app:prompt_parser}

Used in Stage~1 to condense a parsed paper into poster-ready content.

\paragraph{System prompt.} \emph{``You are an expert academic paper analyst. Your task is to read parsed LaTeX paper content and reorganize it into a concise, well-structured JSON format suitable for academic poster generation.''}

\paragraph{User prompt.} Provides the parsed metadata (title, authors, affiliations), abstract, and each section's content (capped at 3,000 characters per section, with subsections capped at 1,500 characters). Instructions:
\begin{enumerate}[leftmargin=*,noitemsep]
  \item Review all sections and determine importance for a poster.
  \item Condense each section to 50--200 words.
  \item Preserve key technical details, results, and conclusions.
  \item Ensure coverage: Title \& Authors, Introduction/Motivation, Method, Results, Conclusion.
  \item Keep content factual---do not add information not in the paper.
  \item Create short section titles (1--4 words).
\end{enumerate}

\paragraph{Output.} JSON with \texttt{meta} (poster\_title, authors, affiliations) and \texttt{sections[]} (title, content, importance: \texttt{high/medium/low}).

\subsubsection{Prompt 4: Poster Planner (PlannerAgent)}
\label{app:prompt_planner}

Used in the baseline (non-template) pipeline to plan poster structure. Not used in the template-filling pipeline where the template's layout is preserved.

\paragraph{System prompt.} \emph{``You are an expert academic poster designer. You plan the structure and layout of academic conference posters.''}

\paragraph{User prompt.} Provides poster dimensions, paper content (sections with importance, first 500 characters each), available figures/tables with captions. If a reference image is provided, the prompt instructs the planner to replicate its column structure, header style, section arrangement, figure placement, and overall flow. Instructions:
\begin{enumerate}[leftmargin=*,noitemsep]
  \item Select 5--8 sections for the poster.
  \item Assign figures/tables to relevant sections.
  \item Decide column layout (matching reference if provided).
  \item Assign content priority (space weight per section).
\end{enumerate}

\paragraph{Output.} JSON with \texttt{layout} (num\_columns, orientation, color\_scheme) and \texttt{poster\_sections[]} (title, content\_source, figure\_id, table\_id, space\_weight, position\_hint).

\subsubsection{Prompt 5: HTML Generator (GeneratorAgent)}
\label{app:prompt_generator}

Used in the baseline pipeline for from-scratch HTML generation. Not used in the template-filling pipeline.

\paragraph{System prompt.} \emph{``You are an expert HTML/CSS developer specializing in academic poster design. You create beautiful, professional conference posters as self-contained HTML files.''}

\paragraph{User prompt.} Provides poster dimensions, the layout plan JSON, and poster section content with figure placeholders (\texttt{FIGURE\_PLACEHOLDER\_N}). The prompt supports three style-guidance modes: (a) image + CSS spec from StyleExtractor; (b) image only; (c) HTML template source. Design requirements include: cohesive color scheme, prominent header, $N$-column grid, clear section headers, readable font sizes (title: 42--48px, headers: 28--32px, body: 18--22px), and CSS Grid/Flexbox layout. 11 critical rules enforce dimension accuracy, placeholder preservation, visual hierarchy, and style matching.

\paragraph{Output.} Complete self-contained HTML.

\subsubsection{Prompt 6: Content Mapper (MappingAgent)}
\label{app:prompt_mapper}

\paragraph{System prompt.} \emph{``You are an expert academic poster designer. Your task is to map the content of a research paper into the slots of an existing poster template. You do NOT design layout or CSS---the template's visual design is fixed. You only decide WHAT TEXT goes into WHICH SLOT.''}

\paragraph{User prompt.} Provides four input blocks:
\begin{enumerate}[leftmargin=*,noitemsep]
  \item \textbf{Paper content}: metadata, abstract, summarized sections (with importance), figures (ID + caption + paths), tables, pre-rendered equations/tables, and optionally raw section content (up to 2,000 chars/section) for additional material.
  \item \textbf{Template slots}: the slot summary generated by the template parser, listing each slot's \texttt{block\_id}, type, label, position, dimensions, font size, and estimated character capacity (\texttt{chars\_per\_line $\times$ max\_lines}).
  \item \textbf{Mapping rules}: 5 action types with detailed constraints (see below).
  \item \textbf{Content distribution rules}: ensuring even distribution across columns.
\end{enumerate}

\paragraph{Mapping action constraints.}
\begin{itemize}[leftmargin=*,noitemsep]
  \item \texttt{replace\_text}: text \textbf{must} fit slot capacity; use bullet points (``\textbullet'') for body text; section headers should be 1--3 words. \textbf{Every text-like slot must use this action.}
  \item \texttt{replace\_image}: only for image-like slots; specify \texttt{figure\_id}; choose figures matching the slot's aspect ratio. \textbf{Use at least 4--6 different paper figures.}
  \item \texttt{replace\_rendered}: for image slots where an equation or table is appropriate; specify \texttt{rendered\_key} (e.g., \texttt{equation\_1}, \texttt{table\_1}).
  \item \texttt{keep}: \textbf{only} for purely decorative elements (borders, dividers, background panels). \textbf{Never} for image slots containing academic content.
  \item \texttt{clear}: \textbf{only} for image-like slots with no matching figure. \textbf{Never} for text-like slots.
\end{itemize}

\paragraph{Positional conventions.} First title slot $\to$ paper title; author slot $\to$ authors/affiliations; largest text slots $\to$ most important sections; top-row $\to$ introduction/motivation; middle $\to$ methods; bottom $\to$ results/conclusion.

\paragraph{Overflow handling.} If more text slots than paper sections, the agent must: split long sections, expand bullet points, add content from abstract/related work, or create ``Key Findings''/``Contributions''/``Future Work'' sections.

\paragraph{Output.} JSON: \texttt{\{slot\_mappings: [\{block\_id, action, new\_text?, figure\_id?, rendered\_key?\}]\}}.

\subsubsection{Prompt 7: Content Adapter (AdapterAgent)}
\label{app:prompt_adapter}

Used after programmatic filling to refine text content length.

\paragraph{System prompt.} \emph{``You are an expert academic poster editor. You refine text content inside an HTML poster template. You MUST NOT change any CSS, layout, positions, colors, font sizes, or DOM structure. You may ONLY edit the text inside \texttt{<div class="content">...</div>} tags.''}

\paragraph{User prompt.} Provides the filled HTML poster (with base64 images stripped and replaced by \texttt{BASE64\_STRIPPED\_N} placeholders to reduce size from $\sim$500KB to $\sim$30KB), the paper's metadata and key sections (up to 300 chars each). The agent's 5 tasks:
\begin{enumerate}[leftmargin=*,noitemsep]
  \item Shorten text that is too long and might overflow its slot.
  \item Expand text that is too short and wastes space.
  \item Format with bullet points and proper line breaks.
  \item Ensure author/affiliation accuracy.
  \item Ensure section headers are clear and concise.
\end{enumerate}

\paragraph{Strict prohibitions (enforced in prompt).}
\begin{itemize}[leftmargin=*,noitemsep]
  \item Do NOT modify \texttt{<style>} tags or any CSS properties.
  \item Do NOT add/remove \texttt{<div>} elements or change \texttt{id}/\texttt{class}/\texttt{style} attributes.
  \item Do NOT change \texttt{<img> src} attributes or \texttt{BASE64\_STRIPPED\_N} placeholders.
  \item ONLY modify text between \texttt{<div class="content">} and \texttt{</div>}.
\end{itemize}

\paragraph{Output.} Complete HTML file with only text content changes. Placeholders are restored to real base64 data URIs after refinement.

\subsubsection{Prompt 8: Visual Reviewer (ReviewerAgent)}
\label{app:prompt_reviewer}

\paragraph{System prompt.} Defines the agent as an expert visual reviewer with expertise in typography, image layout, visual hierarchy, and academic poster aesthetics. Explicitly lists 8 hard constraints:
\begin{itemize}[leftmargin=*,noitemsep]
  \item Must NOT change layout geometry (left, top, width, height).
  \item Must NOT change color schemes, backgrounds, or gradients.
  \item Must NOT change DOM structure (no add/remove elements).
  \item Must NOT hide elements (\texttt{display:none}, \texttt{visibility:hidden}, \texttt{opacity:0}).
  \item Must NOT truncate text or shrink font sizes.
  \item Must NOT add \texttt{overflow:hidden}, \texttt{white-space:nowrap}, or \texttt{text-overflow:ellipsis} to text blocks.
  \item CAN adjust: \texttt{line-height}, \texttt{word-break}, \texttt{overflow-wrap}, \texttt{z-index}, \texttt{padding}.
  \item CAN set \texttt{overflow:hidden} on image blocks only.
\end{itemize}
Critical context: the poster content is \emph{intentionally} different from the original template; content mismatch must not be reported as an issue.

\paragraph{User prompt.} Provides: poster dimensions, slot information summary, previous round fixes (if any), and a structured 5-item checklist:
\begin{enumerate}[leftmargin=*,noitemsep]
  \item \textbf{Text overflow} $\to$ \texttt{word-wrap:break-word}, \texttt{overflow-wrap:break-word}, tighter \texttt{line-height}. Never truncate, never \texttt{overflow:hidden}, never shrink font-size.
  \item \textbf{Text too small} $\to$ increase \texttt{font-size}. Never decrease.
  \item \textbf{Image display} $\to$ \texttt{img::object-fit:contain}, \texttt{img::max-width:100\%}, \texttt{img::max-height:100\%}.
  \item \textbf{Overlapping elements} $\to$ adjust \texttt{z-index} only.
  \item \textbf{Poor readability} $\to$ adjust \texttt{line-height} (not font-size).
\end{enumerate}

\paragraph{Non-issues (must not be reported).} Content mismatch with original template; broken/placeholder images; intentionally empty blocks.

\paragraph{Output format.} JSON:
\begin{verbatim}
{
  "overall_score": <1-10>,
  "issues_found": <int>,
  "review_notes": "<1-2 sentence assessment>",
  "fixes": [{
    "block_id": "blk-XXXX",
    "issue_type": "<text_overflow|text_too_small|
        image_overflow|overlap|readability|...>",
    "description": "<what is wrong>",
    "css_fixes": {"<property>": "<value>"},
    "truncate_text": false
  }]
}
\end{verbatim}

\paragraph{Scoring rubric.}
\begin{itemize}[leftmargin=*,noitemsep]
  \item 9--10: Near perfect---clean, readable, professional, all content fits within blocks.
  \item 7--8: Good---minor issues like slightly tight text, but overall professional.
  \item 5--6: Acceptable---some visible overflow or sizing issues, but poster is usable.
  \item 3--4: Poor---major overflow, unreadable text, or significant layout problems.
  \item 1--2: Unusable---severe rendering failures.
\end{itemize}

\paragraph{CSS fix conventions.} Property names in kebab-case (\texttt{font-size}, not \texttt{fontSize}); values include units (\texttt{16px}, \texttt{1.2}); image-specific fixes use \texttt{img::} prefix (e.g., \texttt{"img::object-fit": "contain"}).

\subsection{Programmatic Filling Details}
\label{app:filling_details}

\paragraph{Text replacement.} For each \texttt{replace\_text} action, the filler locates the target block by regex matching on \texttt{<div id="blk-XXXX">} and replaces the inner HTML of its \texttt{<div class="content">} child. Existing HTML tags within the new text (e.g., \texttt{<b>}, \texttt{<i>}, \texttt{<sub>}, \texttt{<sup>}) are preserved to support inline formatting.

\paragraph{Image replacement.} For \texttt{replace\_image}, the filler resolves the figure ID through the figure registry, attempting multiple path variants (with/without extension, with/without directory prefix such as \texttt{fig/}, \texttt{figures/}, \texttt{images/}) to handle diverse LaTeX path conventions. The resolved base64 data URI replaces the \texttt{src} attribute.

\paragraph{Equation and table rendering.} LaTeX equations and tables are rendered to PNG via MathJax (for inline and display math) and pdfLaTeX (for complex table environments), respectively. Rendered images are stored as base64 data URIs in a separate equation/table registry and injected via the \texttt{replace\_rendered} action.

\paragraph{Safety passes.}
After all mappings are applied, two safety passes run:
\begin{enumerate}[leftmargin=*,noitemsep]
  \item \textbf{Unmapped image clearing}: any image-like slot not covered by the mapping receives \texttt{display:none} to avoid showing template placeholder images from the original poster.
  \item \textbf{Stale asset sweep}: any \texttt{<img>} element whose \texttt{src} still references \texttt{assets/*} (template-local crops) is hidden, preventing residual template content from leaking into the generated poster.
\end{enumerate}

\paragraph{Font size optimization.} After filling, text slot font sizes are dynamically adjusted based on the new content length relative to the slot's capacity. This ensures that the text fills each slot comfortably without overflow or excessive whitespace.

\subsection{CSS Safety Guard Implementation}
\label{app:safety_guards}

The ReviewerAgent's CSS fix application includes a multi-layer safety mechanism to prevent destructive modifications:

\begin{table}[h]
\centering
\caption{CSS safety guard rules. Any VLM-proposed fix matching these conditions is silently blocked.}
\label{tab:safety_guards}
\small
\begin{tabular}{lll}
\toprule
\textbf{Rule} & \textbf{Blocked Property} & \textbf{Blocked Values} \\
\midrule
Element hiding & \texttt{display} & \texttt{none}, \texttt{hidden}, \texttt{0}, \texttt{collapse} \\
Visibility hiding & \texttt{visibility} & \texttt{hidden} \\
Opacity hiding & \texttt{opacity} & \texttt{0} \\
Font shrinking & \texttt{font-size} & Any value $<$ template original \\
Text clipping & \texttt{overflow} (text blocks) & \texttt{hidden}, \texttt{clip} \\
No-wrap & \texttt{white-space} (text blocks) & \texttt{nowrap}, \texttt{pre} \\
Ellipsis & \texttt{text-overflow} (text blocks) & \texttt{ellipsis}, \texttt{clip} \\
\bottomrule
\end{tabular}
\end{table}

For each proposed CSS fix, the guard: (1) checks the property--value pair against the forbidden list; (2) for font-size changes, retrieves the original template font size and rejects any decrease; (3) distinguishes text-like and image-like blocks, applying text-hostile property blocking only to text blocks (image blocks \emph{may} receive \texttt{overflow:hidden} to contain images within their bounding box).

\noindent\begin{minipage}{\textwidth}
\captionsetup{hypcap=false}
\subsection{Post-Training Dynamics}
\label{app:posttraining_dynamics}

Figures~\ref{fig:sft_training_curves} and~\ref{fig:rl_training_curves} summarize the two-stage post-training dynamics. During SFT, training and validation loss decrease as next-token accuracy improves; validation loss and accuracy approach 0.24 and 92.2\%, respectively, as the learning rate decays. During RL, the mean total reward rises from approximately 0.40 to 0.69 over 300 steps. The model-based reward shows the larger increase, while the rule-based reward remains high after an initial rise. Completion length decreases from roughly 5.1k to 4.2k tokens. These curves characterize training-objective optimization rather than held-out poster quality, which is evaluated separately in the main experiments.

\vspace{0.6em}
  \centering
  \includegraphics[width=\textwidth]{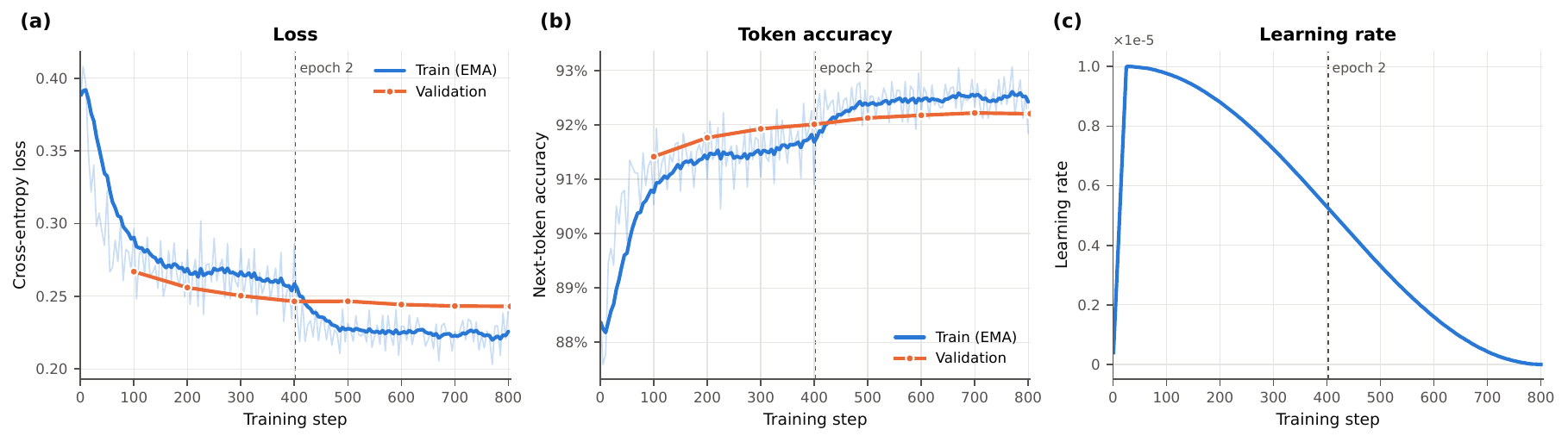}
  \captionof{figure}{Supervised fine-tuning dynamics of Qwen3.8-27B: (a) cross-entropy loss, (b) next-token accuracy, and (c) learning rate. In (a--b), pale blue traces show per-step training measurements, bold blue curves show their exponential moving averages, and orange markers show validation measurements. Dashed lines mark the start of the second epoch.}
  \label{fig:sft_training_curves}
\end{minipage}

\noindent\begin{minipage}{\textwidth}
\captionsetup{hypcap=false}
\centering
  \includegraphics[width=\textwidth]{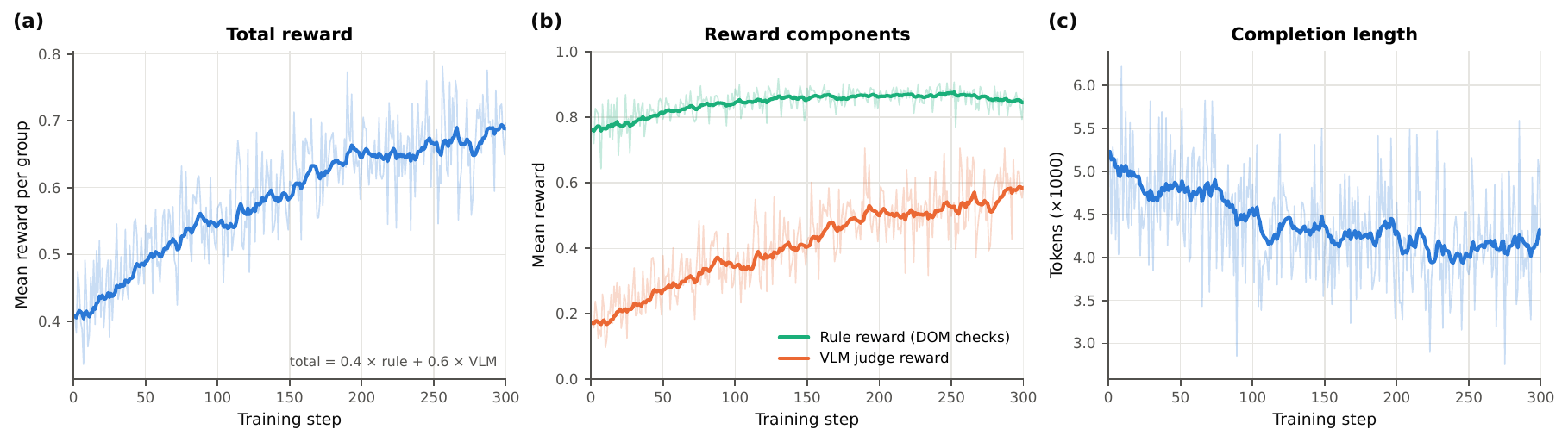}
  \captionof{figure}{Reinforcement-learning dynamics after SFT: (a) mean total reward, $R=0.4R_{\mathrm{rule}}+0.6R_{\mathrm{model}}$; (b) rule-based and model-based reward components; and (c) generated completion length. Light traces show per-step measurements, while bold curves show smoothed trends.}
  \label{fig:rl_training_curves}
\end{minipage}

\noindent\begin{minipage}{\textwidth}
\captionsetup{hypcap=false}
\subsection{Ablation on RL Rewards}
\label{app:rl_reward_ablation}

Table~\ref{tab:reward_ablation} compares the rule-based reward, the model-based reward, and their combination.
The combined configuration outperforms both single-reward variants on LPIPS, DISTS, FID, and FDiD.
Compared with using the model-based reward alone, it reduces FID from 25.58 to 23.42 and FDiD from 313.34 to 306.94, while DINOv2-CosDist decreases slightly from 0.2784 to 0.2735.
The combined objective thus improves distributional human-likeness and two of the three diversity measures.

\centering
\footnotesize
\setlength{\tabcolsep}{3pt}
\captionof{table}{Ablation of RL reward components on Qwen3.8-27B. Rule-based and Model-based denote DOM-rule and VLM-based rewards, respectively. Best results are in \textbf{bold}; second-best results are \underline{underlined}.}
\label{tab:reward_ablation}
\resizebox{\columnwidth}{!}{%
\begin{tabular}{l |cc| ccc cc}
\toprule
\multirow{2}{*}{\textbf{Setting}}
  & \multicolumn{2}{c|}{\textbf{Rewards}}
  & \multicolumn{3}{c}{\textbf{Diversity}}
  & \multicolumn{2}{c}{\textbf{Human-likeness}} \\
\cmidrule(lr){2-3} \cmidrule(lr){4-6} \cmidrule(lr){7-8}
  & \small Rule-based & \small Model-based
  & \small{LPIPS$\uparrow$} & \small DINOv2-CosDist$\uparrow$ & \small DISTS$\uparrow$
  & \small FID$\downarrow$ & \small FDiD$\downarrow$ \\
\midrule
\small{Base} & \small -- & \small --
  & \small{0.6284} & \small0.2579 & \small0.2680 & \small32.44 & \small465.00 \\
\small SFT & \small -- & \small --
  & \small 0.6599 & \small \textbf{0.2784} & \small \underline{0.2892} & \small 28.00 & \small 326.24 \\
\midrule
\small RL  & \small \textcolor{darkgreen}{\cmark} & \small \textcolor{red}{\xmark}
  & \small 0.6554 & \small 0.2733 & \small 0.2834 & \small 27.63 & \small 323.61 \\
\small RL  & \small \textcolor{red}{\xmark} & \small \textcolor{darkgreen}{\cmark}
  & \small \underline{0.6735} & \small \textbf{0.2784} & \small 0.2867 & \small \underline{25.58} & \small \underline{313.34} \\
\cellcolor{lavender}\small RL (ours)
  & \cellcolor{lavender}\small \textcolor{darkgreen}{\cmark}
  & \cellcolor{lavender}\small \textcolor{darkgreen}{\cmark}
  & \cellcolor{lavender}\small \textbf{0.6925}
  & \cellcolor{lavender}\small \underline{0.2735}
  & \cellcolor{lavender}\small \textbf{0.2944}
  & \cellcolor{lavender}\small \textbf{23.42}
  & \cellcolor{lavender}\small \textbf{306.94} \\
\bottomrule
\end{tabular}%
}
\end{minipage}

\section{Evaluation Details}
\label{app:eval}

\subsection{Metrics}

\subsubsection{Diversity}

Given the generated poster set $\{\hat{P}_i\}_{i=1}^{N}$ produced by a method on the $N$-sample test set, we compute each diversity metric pairwise over all $\binom{N}{2}$ pairs and report the average. Higher pairwise distances indicate more structurally and perceptually varied outputs across different input papers.

\paragraph{LPIPS~\citep{Zhang_2018_CVPR}.}
Learned Perceptual Image Patch Similarity measures perceptual distance via weighted $\ell_2$ distance between deep feature activations of a pretrained AlexNet. Higher mean pairwise LPIPS indicates that posters look more visually distinct to human perception across the generated set.

\paragraph{DINOv2-CosDist.}
We extract a global feature embedding from each poster using the pretrained DINOv2~\citep{oquab2023dinov2} ViT-L/14 encoder and compute the cosine distance ($1 - \cos$) between every pair of embeddings. As DINOv2 is trained via self-supervision to capture high-level semantic and stylistic structure, larger DINOv2-CosDist values indicate that the generated posters span a wider range of design styles and visual themes in the model's semantic space, providing a strong indicator of stylistic diversity beyond pixel-level variation.

\paragraph{DISTS~\citep{ding2021dists}.}
Deep Image Structure and Texture Similarity extracts multi-scale VGG features and measures both structural and textural dissimilarity in a distortion-tolerant manner. Higher mean pairwise DISTS indicates greater deep-feature diversity across the generated set.

\subsubsection{Human-Likeness}
To measure the distributional gap between generated posters and real human-created designs, we compute the Fréchet distance in different feature spaces. Given the feature representations of real posters (reference distribution $r$) and generated posters (generated distribution $g$), the Fréchet distance $d^2$ is defined as:
\begin{equation}
    d^2 = \lVert\mu_r - \mu_g\rVert^2 + \text{Tr}\Bigl(\Sigma_r + \Sigma_g - 2\sqrt{\Sigma_r \Sigma_g}\Bigr)
\end{equation}
where $\mu$ and $\Sigma$ denote the mean vector and covariance matrix of the respective feature distributions. To capture different aspects of poster fidelity, we employ two metric variants by changing the underlying feature extractor:

\begin{itemize}[leftmargin=*]
    \item \textbf{Fréchet Inception Distance (FID)~\citep{heusel2017gans}:} Extracts 2048-dimensional features from the \texttt{pool3} layer of an ImageNet-pretrained InceptionV3~\citep{szegedy2016inception}. FID captures general image synthesis quality but is relatively insensitive to document-specific layouts.
    \item \textbf{Fréchet DINOv2 Distance (FDiD):} Extracts 768-dimensional features from the \texttt{pooler\_output} of a self-supervised \texttt{dinov2-base} model~\citep{oquab2023dinov2}. By utilizing self-supervised vision transformer representations, FDiD is highly sensitive to visual structure, spatial arrangement, and layout geometry.
\end{itemize}

Lower values across both metrics indicate that the generated posters are distributionally closer to real human designs.

\end{document}